\documentclass{article}

\usepackage[preprint]{neurips_2026}
\usepackage{amssymb}

\usepackage[utf8]{inputenc} 
\usepackage[T1]{fontenc}    
\usepackage{hyperref}       
\usepackage{url}            
\usepackage{booktabs}       
\usepackage{amsfonts}       
\usepackage{nicefrac}       

\usepackage{amsmath}
\usepackage{amssymb}
\usepackage{mathtools}
\usepackage{amsthm}
\usepackage{bbm}
\usepackage{multirow}
\usepackage{pgfplots}
\usepackage{pgfplotstable}
\usepackage[table]{xcolor}
\usepackage{booktabs}

\usepackage{multirow}
\definecolor{smoothbg}{RGB}{235,242,250}
\pgfplotsset{compat=1.18}
\usepgfplotslibrary{groupplots}

\usepackage{microtype}      
\usepackage{xcolor}         
\theoremstyle{plain}
\newtheorem{theorem}{Theorem}[section]

\theoremstyle{definition}

\theoremstyle{remark}

\newcommand{\citeme}[1]{\red{[XX]}}
\newcommand{\refme}[1]{\red{(XX)}}

\newcommand{\real}{\mathbb{R}}

\newcommand{\defn}{\mathrel{:=}}

\newcommand{\cX}{\mathcal{X}}
\newcommand{\cY}{\mathcal{Y}}

\newcommand{\vc}{\mathbf{c}}

\newcommand{\vx}{\mathbf{x}}

\newcommand{\vz}{\mathbf{z}}

\makeatletter
\newcommand*\bdot{\mathpalette\bdot@{.7}}
\newcommand*\bdot@[2]{\mathbin{\vcenter{\hbox{\scalebox{#2}{$\m@th#1\bullet$}}}}}
\makeatother

\makeatletter
\DeclareRobustCommand\onedot{\futurelet\@let@token\@onedot}
\def\@onedot{\ifx\@let@token.\else.\null\fi\xspace}

\makeatother

\usepackage{algorithm}
\usepackage{algorithmic}
\usepackage{subcaption}
\usepackage{tikz}
\usetikzlibrary{shapes.geometric, positioning, calc, arrows.meta, backgrounds, fit, patterns, matrix}
\usepackage{graphicx}

\usepackage{pgfplots}
\usepackage{subcaption}
\usepackage{pgfplotstable}
\usepgfplotslibrary{colormaps}
\pgfplotsset{compat=1.18}

\newcommand{\takeaway}[1]{\begin{center}
\colorbox{white!90!blue}{\begin{minipage}{0.95\textwidth} \emph{#1}\end{minipage}}
\end{center}}

\title{Interpretable but Fragile? Robustness of Concept Bottlenecks under Geometric-Semantic Perturbations}

\author{
   Hanwei Zhang \\
   Department of Computer Science\\
  Saarland University\\ Saarbrücken, Germany \\
 \texttt{zhang@depend.uni-saarland.de} \\
 \And
 Tianma Hu \\
 School of Computer Science and Engineering\\
 Tianjin University of Technology \\
Tianjin, China \\
\texttt{hutianma15299@stud.tjut.edu.cn} \\
 \AND
 Gaojie Jin \\
 Department of Artificial Intelligence \& \\Institute of Artificial Intelligence and Brain Sciences \\University of Macau\\
 Macau SAR, China \\
\texttt{gaojiejin@um.edu.mo} \\
 \And
 Xu Cheng\\
 School of Computer Science and Engineering\\
 Tianjin University of Technology \\
Tianjin, China \\
\texttt{xu.cheng@ieee.org} \\
 \And
Ronghui Mu\thanks{corresponding author} \\
 Department of Computer Science\\University of Exeter \\
Exeter,UK \\
\texttt{r.mu2@exeter.ac.uk} \\
 }

\begin{document}

\maketitle

\begin{abstract}

Concept Bottleneck Models (CBMs) are designed to provide interpretable intermediate representations, yet how such bottlenecks affect robustness remains unclear, with existing studies reporting mixed and sometimes contradictory findings. We argue that these discrepancies arise from conflating different robustness notions and perturbation regimes, rather than from fundamental disagreements about CBMs themselves.
To disentangle these factors, we introduce a generator-based evaluation framework that enables controlled comparisons between standard classifiers and CBMs  under two distinct perturbation types: continuous geometric perturbations in latent space and discrete semantic interventions in concept space. We further evaluate a prototype-based interpretable model, PixPNet, showing that the same perturbation-to-prediction evaluation and certification pipeline extends beyond concept bottlenecks without requiring alignment between heterogeneous internal representations.
Within this framework, we evaluate robustness both empirically, via prediction and concept-level sensitivity metrics, and certifiably, using randomized smoothing in latent and concept spaces. 
Across experiments on CUB and RIVAL10 variants, we reconcile previously conflicting findings by clarifying when, and in what sense, concept bottlenecks do or do not improve robustness. By further analyzing robustness under varying task conditions, including class semantic similarity and concept vocabulary size, we show that interpretability does not inherently confer robustness. Instead, concept bottlenecks shift where and how sensitivity manifests, revealing a nuanced interpretability robustness trade off that depends critically on the perturbation regime and task structure.
Together, our results show that interpretability and robustness are distinct objectives: interpretable intermediate representations do not uniformly improve robustness, but instead redistribute sensitivity across perturbation spaces and model families.
\end{abstract}

\section{Introduction}

Concept Bottleneck Models (CBMs)~\citep{koh2020concept} are a prominent paradigm of interpretable learning, explicitly structuring predictions around human‑interpretable concepts. By exposing intermediate concept representations, CBMs provide semantically grounded explanations, support targeted interventions, and offer a modular design that extends naturally in vision and multimodal settings. These properties make CBMs attractive, particularly in high‑stakes and regulated domains such as those governed by the EU AI Act~\citep{hermanns2024ai,AIAct}, where interpretability is often believed to promote robustness. However, the robustness implications of introducing a concept bottleneck remain underexplored, and existing evidence is mixed.

\begin{figure*}[t]
    \centering
    
    \begin{subfigure}[b]{0.3\textwidth}
        \centering
        \includegraphics[width=\textwidth]{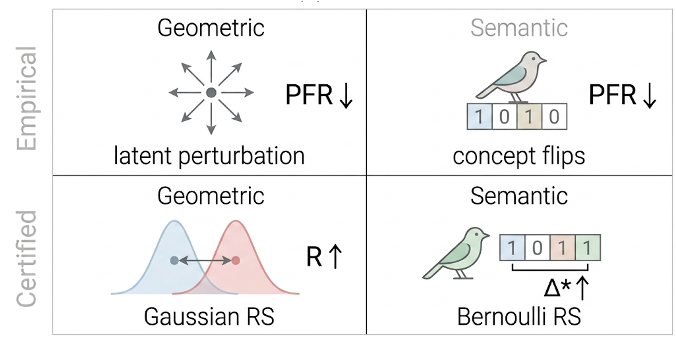}
        \caption{Evaluation framework}
        \label{fig:eval_axes}
    \end{subfigure}
    \hfill
    \begin{subfigure}[b]{0.4\textwidth}
        \centering
        \includegraphics[width=\textwidth]{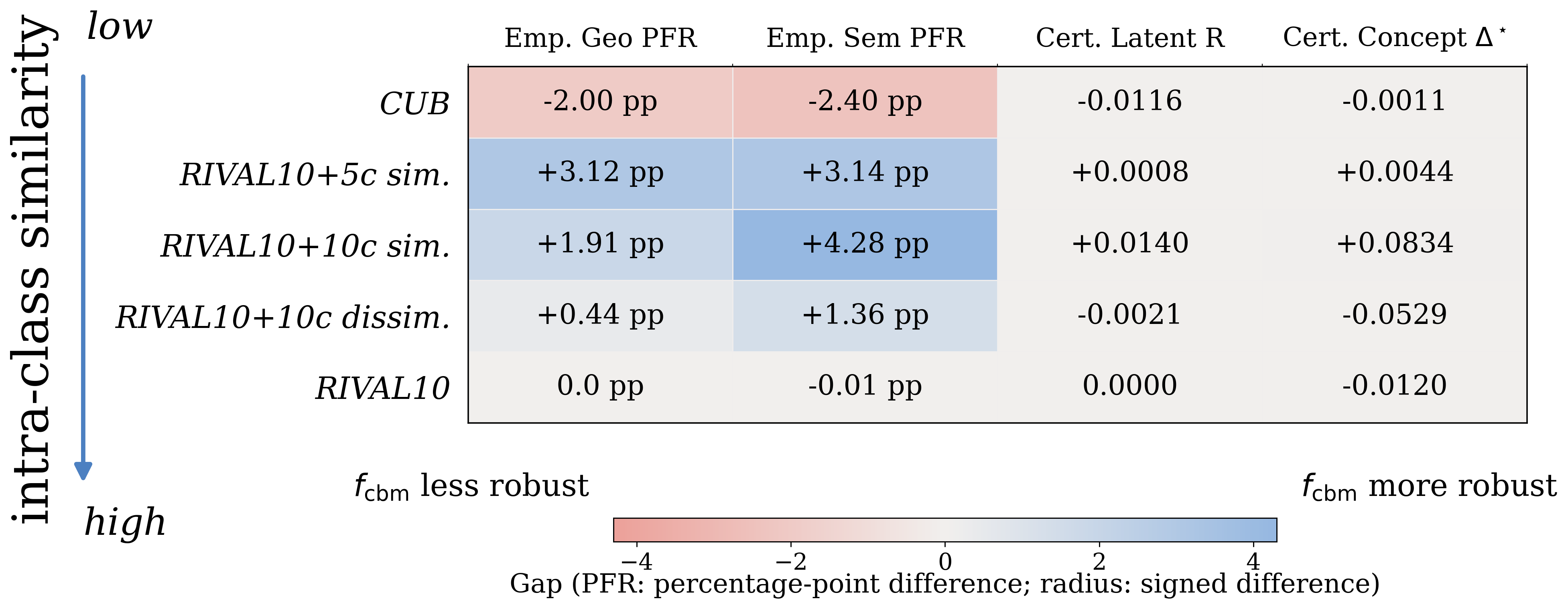}
        \caption{Robustness gap summary}
        \label{fig:gap_summary}
    \end{subfigure}
    \hfill
    \begin{subfigure}[b]{0.27\textwidth}
        \centering
        \includegraphics[width=0.6\textwidth]{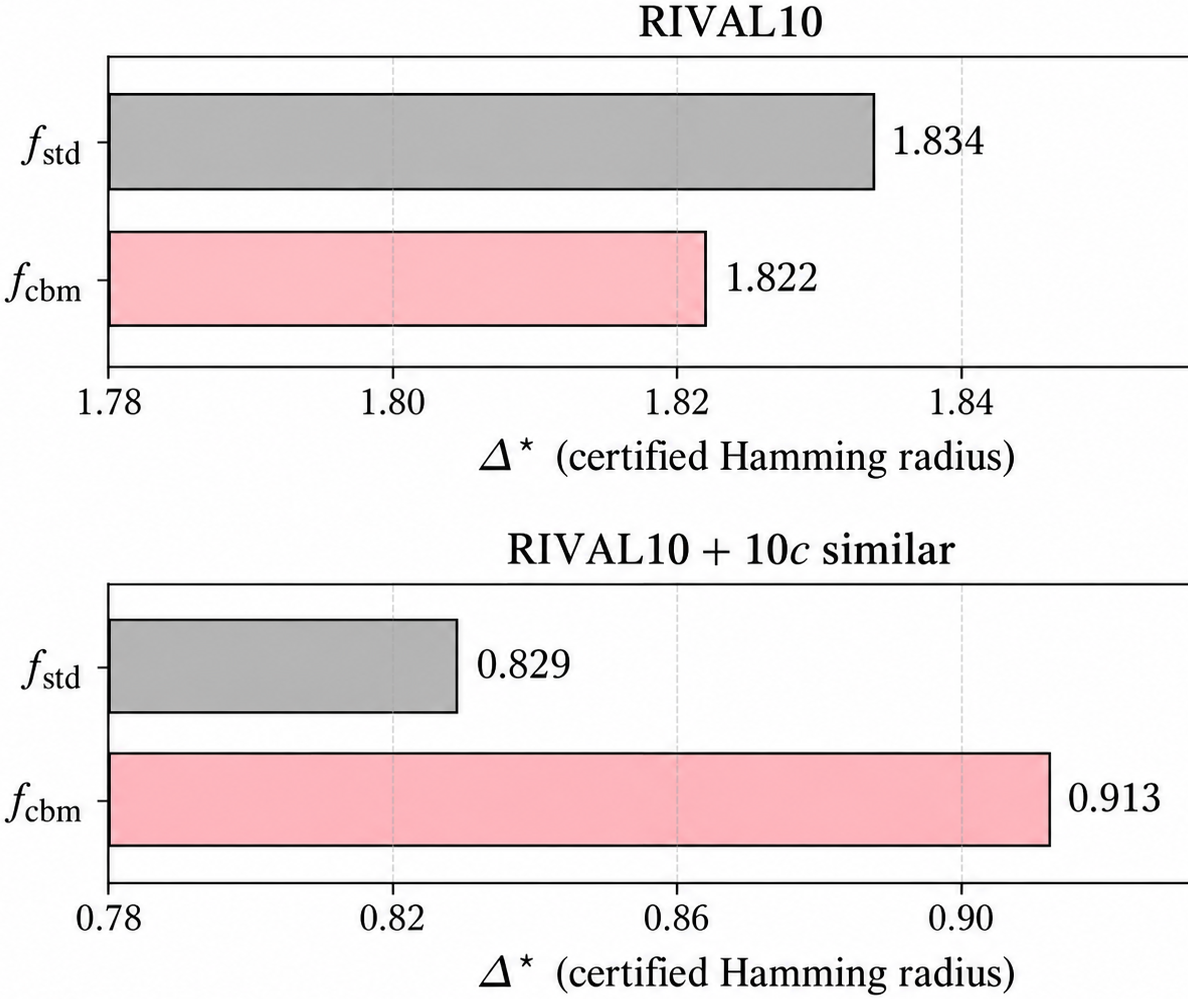}
        \caption{Concrete contrast}
        \label{fig:concrete_contrast}
    \end{subfigure}

    \vspace{1em} %

    \caption{ \textbf{Evaluation framework and representative results.}
(a) Evaluation robustness across geometric \emph{vs.} semantic perturbations and empirical \emph{vs.}
certified criteria. 
(b) The heatmap reports representative $f_{cbm}$--$f_{std}$ robustness gaps using
$\epsilon=0.3$, $\tau=5$, $\sigma=0.10$, and $\rho=0.10$ for the four columns, respectively.
All gaps are oriented so that positive values indicate greater CBM robustness. Detailed computations in Appendix \ref{app:experiments_fig1}
(c) Certified concept-radius examples at $\rho=0.10$ show opposite outcomes across task settings. }
    \label{fig:teaser}
\end{figure*}

Prior work reaches conflicting conclusions regarding the robustness of CBMs.
\citet{rasheed2024exploring} report robustness gains from concept bottlenecks, attributing them to the removal of non‑essential variation, with robustness evaluated at the level of final class predictions. 
In contrast, \citet{sinha2023understanding} argues that concept bottlenecks can reduce robustness: concept representations themselves are often unstable under perturbations, and exposing an explicit concept interface introduces additional attack surfaces. 
As illustrated in \autoref{fig:teaser}(c), we observe this tension directly: under different task settings of the same dataset, evaluating the same standard classifier and its concept bottleneck counterpart can lead to opposite conclusions about whether concept bottlenecks improve robustness.

At first glance, these findings appear irreconcilable. We argue that they instead stem from different notions of robustness and perturbation regimes. In practice, robustness is still most often assessed using small pixel‑level perturbations, which lack semantic meaning and fail to reflect changes in underlying concepts. Whether a concept bottleneck improves or degrades robustness depends on \emph{where robustness is measured} (class \emph{vs.} concept level), \emph{how perturbations are applied}, and \emph{the task conditions} under which the model operates. Making these distinctions explicit is key to understanding the role of concept bottlenecks in robust learning.
This leads to the central question of this work:
\begin{quote}
\textit{How does introducing an interpretable intermediate representation reshape robustness relative to a standard classifier, and how does this depend on the perturbation space and robustness criterion?}
\end{quote}
To study this systematically, we adopt a concept bottleneck generator~\citep{kulkarni2025interpretable} that maps latent representations to an explicit concept space, enabling controlled concept‑level interventions before image generation. The generator is frozen and shared across classifiers, serving only as an experimental instrument.
This design allows us to disentangle two fundamentally different types of perturbations: 
(i) \emph{geometric perturbations}, which induce continuous variations in the latent embedding space, and 
(ii) \emph{semantic perturbations}, which directly modify individual concepts.
Within this framework, we evaluate robustness \emph{empirically} by measuring changes in class predictions and intermediate representations, and \emph{certifiably} by deriving guarantees via randomized smoothing. Standard models and their corresponding CBMs share the same backbone, ensuring a fair comparison.

Our analysis focuses on generator‑based perturbations to enable controlled evaluation of the robustness induced by concept bottlenecks, without claiming universal robustness guarantees.
Our results reveal distinct robustness profiles across perturbation regimes: concept bottlenecks can suppress prediction sensitivity in some settings while introducing additional vulnerabilities in others. In particular, we find that the effect of the bottleneck is strongly influenced by the semantic similarity between classes, and the number and relevance of concepts used by the model. These findings provide a unifying explanation for previously conflicting results and clarify the conditions under which concept bottlenecks act as beneficial regularizers versus when they introduce fragility.

Overall, our contributions are threefold:

(i) \textbf{A unified robustness perspective for concept bottlenecks.}
We reconcile previously conflicting findings on CBM robustness by showing that they often arise from conflating different perturbation spaces and robustness criteria. By distinguishing prediction-level and concept-level sensitivity under geometric and semantic perturbations, as well as empirical, certified, and adversarial robustness, we clarify when concept bottlenecks suppress sensitivity and when they instead introduce additional vulnerabilities. Our results emphasize that interpretability and robustness are distinct objectives, and that concept bottlenecks reshape robustness rather than uniformly improving it.

(ii) \textbf{A controlled evaluation and certification framework for geometric and semantic robustness.}
We introduce a generator-based framework (Figure~\ref{fig:teaser}(a)) that separates continuous latent-space geometric perturbations from discrete concept-level semantic interventions. The framework enables matched comparisons under a common perturbation-to-prediction interface, together with empirical robustness metrics and perturbation-specific randomized-smoothing certificates in latent and concept spaces. Importantly, the evaluation interface does not require different classifier architectures to share or align their internal representations.

(iii) \textbf{Empirical and certified analysis of when robustness gains emerge.}
Through extensive experiments on CUB-200-2011 and RIVAL-10 variants, we show that concept bottlenecks do not provide an inherent robustness advantage (\emph{e.g.}, Figure~\ref{fig:teaser}(b)). Instead, robustness depends strongly on the perturbation regime, robustness criterion, class semantic similarity, and concept vocabulary design. These findings provide practical guidance for designing CBMs that better balance interpretability and robustness. We further validate the broader applicability of the evaluation pipeline on PixPNet, a prototype-based interpretable model, showing that the same generator-defined evaluation and certification procedure can be applied beyond CBMs without requiring alignment between generator concepts and model-specific prototypes.

\section{Related Work}

\paragraph{Robustness of Concept Bottlenecks.}
Early work on interpretable models highlighted connections between interpretability and robustness, with Self‑Explaining Neural Networks (SENN) enforcing robustness as the stability of explanations under small perturbations~\citep{sawada2022c}. More recent studies directly examine robustness in Concept Bottleneck Models (CBMs). \citet{sinha2023understanding} show that interpretability alone does not guarantee robustness, demonstrating that concept representations can be fragile under adversarial perturbations and that explicit concept interfaces introduce new attack surfaces, revealing a tension between concept‑level stability and adversarial robustness. In contrast, \citet{rasheed2024exploring} find that CBMs, particularly sequentially trained variants, can achieve higher adversarial accuracy than end‑to‑end models under weak to moderate attacks, suggesting that conceptual bottlenecks may act as an information‑filtering regularizer at the prediction level. Related observations beyond CBMs further indicate that highly interpretable representations can remain non‑robust~\citep{li2026evaluating}, while recent extensions such as language‑guided and flexible CBMs improve scalability and adaptability without addressing robustness explicitly~\citep{yu2025language,du2026flexible}. Overall, prior work reports seemingly conflicting conclusions, which we argue arise from differences in robustness definitions, perturbation models, and task settings rather than fundamental inconsistencies.
\paragraph{Randomized Smoothing for Certified Robustness.}
Randomized smoothing provides a general framework for certifying robustness by constructing a smoothed classifier through noise injection and aggregation, yielding probabilistic guarantees against bounded perturbations~\citep{cohen2019certified}. While most work applies smoothing in pixel space, recent studies have explored extensions to learned representations or structured perturbations~\citep{hao2022gsmooth,wang2026clucert,mu2023certified,mu2024reward}. In this work, randomized smoothing is used primarily as a
model-agnostic certification and comparison tool. The resulting smoothed
predictor can also improve empirical robust accuracy in some adversarial
settings, but defense is not the primary objective of our analysis.

\paragraph{Interpretable intermediate architectures beyond CBMs.}
Interpretable intermediate representations are also exposed by post-hoc concept methods~\citep{yuksekgonul2022post} and prototype-based architectures~\citep{zhang2021prototype,carmichael2024pixel,sicre2023dp}. PCBM~\citep{yuksekgonul2022post} constructs concept representations using CAV-style directions, while prototype-based models such as PixPNet~\citep{carmichael2024pixel} ground predictions in localized parts. Although these models expose different internal representations, our evaluation does not require these representations to be aligned: the generator defines a common external perturbation interface, and robustness is measured at the resulting prediction level.


\section{Methodology}

\subsection{Perturbation Models}
Let $g:\real^d\rightarrow\cX$ denote a pretrained generator
that maps a latent noise vector $\vz\in\real^d$ to an image
$\vx=g(\vz)$. We assume that $g$ admits a decomposition
$g=g_2\circ g_1$, where $g_1$ maps the latent code to an
intermediate representation and $g_2$ maps this
representation to the image space.
Following concept bottleneck generators, we introduce an
encoder--decoder pair $(E,D)$ at an intermediate generator
layer. The encoder maps the intermediate features
$g_1(\vz)$ to a concept representation
$\vc=E(g_1(\vz))\in\real^{2K}$, containing $K$ concepts,
each represented by a pair of state coordinates
$(c_i^+,c_i^-)$.
The decoder maps this representation back to the generator
feature space, yielding
$g(\vz)=g_2(D(E(g_1(\vz))))$.
Each concept corresponds to a semantically meaningful
attribute, with its binary state indicating the presence
or absence of the corresponding attribute.
This representation allows us to distinguish two
fundamentally different perturbation regimes:
\paragraph{Geometric perturbations} are applied in latent space by adding noise to the latent code, $\tilde{\vz} = \vz + \mathbf{\delta}, \|\mathbf{\delta}\|_2 \leq \epsilon$, which induces a continuous variation along the generator manifold and results in a geometrically perturbed image $\tilde{\vx} = g(\tilde{\vz})$. 
\paragraph{Semantic perturbations} are applied directly in concept space by modifying the concept vector $\vc$. We uniformly sample a subset $S \subseteq \{1, \cdots, K\}$ of size $|S| = \tau$ and, for each $i\in S$, swap the binary representation of corresponding concept from $[c_i^+,c_i^-]$ to $[c_i^-,c_i^+]$, simulating changes in high-level semantic attributes while preserving the underlying geometric structure. To detect concept changes under perturbation,
we train binary concept classifiers
$C(\vc)=(C_1(\vc),\ldots,C_K(\vc))
\in\{0,1\}^K$
that map the concept representation to binary
concept states. We define the concept-flip indicator as
\begin{equation}
\phi_i(\vc,\tilde{\vc})
:=
\mathbb{1}\left[
C_i(\vc)\neq C_i(\tilde{\vc})
\right].
\end{equation}

\subsection{Statistical Robustness Analysis}

Let $f: \cX \to \cY$ denote a classifier operating on generated images. We consider two classes of models: \emph{standard classifier} $f_{std}$ directly maps the input image to a label; \emph{concept bottleneck model} is defined as a tuple $f_{cbm} \defn (r,h)$, where $r: \cX \to \real^K$ predicts a concept representation from image and $h: \real^K \to \cY$ maps concepts to the final prediction.
This setup enables a controlled comparison of how standard and concept‑based classifiers respond to geometric perturbations in latent space and semantic perturbations in concept space. Further methodological details are provided in Appendix.

\paragraph{Empirical Robustness.}

To quantify empirical robustness under geometric and semantic perturbations, we measure changes at the concept and prediction levels using the following metrics:

   \emph{1. Concept Flip Rate (CFR)} measures the average
number of concepts whose predicted states change
under perturbation:
\begin{equation}
\mathrm{CFR}
:=
\mathbb{E}_{(\vc,\tilde{\vc})}
\left[
\sum_{i=1}^{K}
\phi_i(\vc,\tilde{\vc})
\right].
\end{equation}
   \emph{2. Prediction Flip Rate (PFR)} measures the probability that the classifier’s predicted label changes under perturbation:
    \[ \mathrm{PFR}(f)
    \;\defn\; \mathbbm{E}_{\vx} \left[
    \mathbbm{1}
    \bigl[
    f(\mathbf{x}) \neq f(\tilde{\mathbf{x}})
    \bigr] \right];
    \]
    \emph{3. Decision Function Sensitivity (DFS)} quantifies the sensitivity of a concept bottleneck classifier $f_{\mathrm{cbm}}$ by measuring the change in its internal concept representation:
    \[
    \mathrm{DFS}
    \;\defn\; \mathbbm{E}_{\vz} \left[
    \left\|
    r\bigl(g(\mathbf{z})\bigr)
    -
    r\bigl(g(\tilde{\mathbf{z}})\bigr)
    \right\|_2 \right].
    \]

CFR and DFS characterize concept‑level robustness, capturing both true concept flips induced by the generator and the sensitivity of the concept bottleneck classifier, while PFR measures robustness at the prediction level. Together, these metrics allow us to study how concept instability propagates to model predictions.

\paragraph{Certified Robustness.}

We use randomized smoothing to certify robustness in the two perturbation spaces
induced by the generator.

\paragraph{Latent-space smoothing.}
For geometric robustness, we apply Gaussian smoothing in latent space:
\begin{equation}
\hat f_{\mathrm G}(\vz)
=
\arg\max_{y\in\mathcal Y}
\mathbb P_{\delta\sim\mathcal N(0,\sigma^2 I)}
\!\left[f(g(\vz+\delta))=y\right].
\label{eq:main_latent_smoothed}
\end{equation}

\begin{theorem}[Latent-space Gaussian certificate~\citep{cohen2019certified}]
\label{thm:latent_cert_main}
Let $A$ and $B$ be the top and runner-up classes of $\hat f_{\mathrm G}$ with
probabilities $p_A>p_B$. Then $\hat f_{\mathrm G}$ is constant for all latent
perturbations $\|\epsilon\|_2\le R_{\mathrm G}$, where
\begin{equation}
R_{\mathrm G}
=
\frac{\sigma}{2}
\left(\Phi^{-1}(p_A)-\Phi^{-1}(p_B)\right).
\label{eq:main_latent_radius}
\end{equation}
\end{theorem}

\paragraph{Concept-space smoothing.}
For semantic robustness, we sample independent concept flips
$\eta_i\sim\mathrm{Bernoulli}(\rho)$ and write $\mathcal F(\vc,\eta)$ for the
flipped concept representation. The smoothed classifier is
\begin{equation}
\hat f_{\mathrm C}(\vc)
=
\arg\max_{y\in\mathcal Y}
\mathbb P_{\eta\sim\mathrm{Bernoulli}(\rho)^K}
\!\left[
f\!\left(g_2(D(\mathcal F(\vc,\eta)))\right)=y
\right].
\label{eq:main_concept_smoothed}
\end{equation}

\begin{theorem}[Concept-space $\ell_0$ certificate~\citep{lee2019tight}]
\label{thm:concept_cert_main}
Let $y^\star$ be the top class of $\hat f_{\mathrm C}$, and let
$\underline p_{y^\star}$ and $\overline p_y$ be conservative confidence bounds
on the top and competing class probabilities. For any concept-flip budget
$\Delta$, if
\begin{equation}
\mathcal L_\Delta(\underline p_{y^\star})
>
\mathcal U_\Delta(\overline p_y),
\qquad
\forall y\neq y^\star,
\label{eq:main_concept_condition}
\end{equation}
then, with probability at least $1-\alpha$ over Monte Carlo estimation,
$\hat f_{\mathrm C}(\tilde{\vc})=y^\star$ for all
$\|\mathcal C(\tilde{\vc})-\mathcal C(\vc)\|_0\le\Delta$.
\end{theorem}

We report the largest certified semantic radius
\begin{equation}
\Delta^\star(\vc)
=
\max\left\{
\Delta:
\mathcal L_\Delta(\underline p_{y^\star})
>
\mathcal U_\Delta(\overline p_y),\ \forall y\neq y^\star
\right\}.
\label{eq:main_concept_radius}
\end{equation}
All confidence bounds use Clopper--Pearson intervals; full transfer-function
definitions, proofs, and algorithms are in Appendix.

\section{Experiments}

In this section, we empirically and certifiably evaluate the robustness of $f_{std}$ and $f_{cbm}$ under controlled geometric and semantic perturbations. Due to page limitations, we report a subset of the experimental results in the main paper; the complete results, including additional statistical analyses and backbone comparisons, are provided in \autoref{app:experiments}. Across perturbation regimes, task conditions, and evaluation metrics, our experiments reveal a consistent pattern:
\emph{Concept bottlenecks do not induce monotonic robustness improvements; instead, they shift where and how sensitivity manifests.}

\subsection{Experimental Setup}

We conduct experiments on two datasets with rich concept annotations: CUB‑200‑2011 and RIVAL‑10.
CUB‑200‑2011~\citep{WahCUB_200_2011} is a fine‑grained bird classification benchmark consisting of 11,788 images from 200 species, each annotated with 112 expert‑defined visual concepts.\footnote{\scriptsize\url{https://www.vision.caltech.edu/datasets/cub_200_2011/}}
RIVAL‑10~\citep{moayeri2022comprehensive} aligns CIFAR‑10~\citep{krizhevsky2009learning} classes with ImageNet~\citep{krizhevsky2012imagenet} categories and contains approximately 26,000 images annotated with 18 visual attributes and corresponding segmentations.

For CUB‑200‑2011, we select $K=15$ concepts by default based on empirical frequency. We adopt a post‑hoc concept bottleneck autoencoder (CB‑AE)~\citep{kulkarni2025interpretable} built on StyleGAN~\citep{karras2020analyzing} as the generator, and train it on CUB‑200‑2011 to induce an explicit concept space aligned with these concepts. 
For RIVAL‑10, models are trained on the official training split of 21,098 images and evaluated on the test split of 5,286 images, with all reported results computed on the test set. We adopt BigGAN~\citep{brock2018large} as the base image generator and treat all 18 annotated attributes as concepts by default.
For classification, all compared classifiers share the same pretrained backbone to ensure a fair and controlled comparison.
For both datasets, we use a ResNet‑50~\citep{he2016deep} for the standard classifier $f_{std}$ and a ResNet‑50 version of the post‑hoc concept bottleneck model (PCBM)~\citep{yuksekgonul2022post} for the concept‑based classifier $f_{cbm}$. 

We restrict evaluation to generated inputs on which $f_{std}$ and $f_{cbm}$ agree,
denoted by $\mathcal{S}_{\mathrm{agree}}$, and use the shared prediction as a reference label.

\subsubsection{Empirical robustness results}

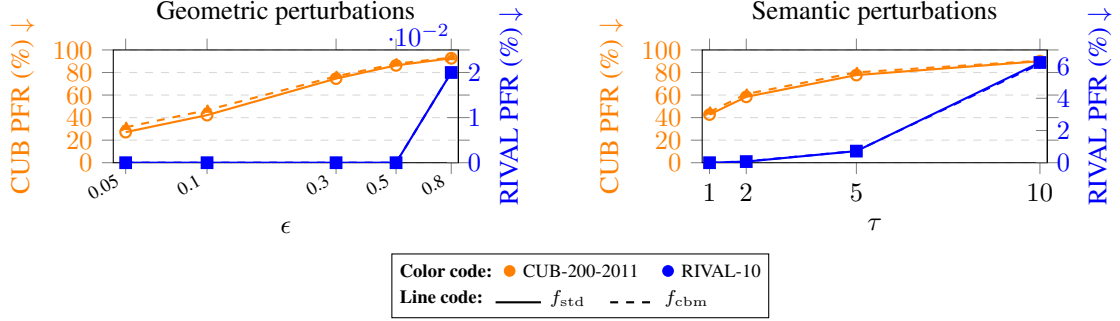
\begin{figure}[t] 
\centering \begin{tikzpicture} 
\begin{axis}[ name=geo, width=0.44\linewidth, height=0.22\linewidth, title={Geometric perturbations}, xlabel={$\epsilon$}, ylabel={CUB PFR (\%) $\downarrow$}, ylabel style={orange}, yticklabel style={orange}, 
xmode=log,xmin=0.045, xmax=0.85, ymin=0,ymax=100, xtick={0.05,0.1,0.3,0.5,0.8},
  xticklabels={0.05,0.1,0.3,0.5,0.8},xticklabel style={font=\scriptsize, rotate=30, anchor=east},ymajorgrids, grid style={dashed,gray!30}, tick align=outside, ] 
\addplot[ orange, thick, mark=o ] coordinates { (0.05,27.1) (0.1,42.2) (0.3,74.6) (0.5,86.3) (0.8,92.6) }; 
\addplot[ orange, thick, dashed, mark=triangle* ] coordinates { (0.05,31.4) (0.1,46.6) (0.3,76.6) (0.5,87.8) (0.8,93.2) }; 
\end{axis} 
\begin{axis}[
  at={(geo.south west)}, anchor=south west,
  width=0.44\linewidth, height=0.22\linewidth,
  xmode=log,
  xmin=0.045, xmax=0.85,
  ymin=0, ymax=0.025,
  ylabel={RIVAL PFR (\%) $\downarrow$},
  ylabel style={blue}, yticklabel style={blue},
  axis y line*=right, axis x line=none,
]\addplot[ blue, thick, mark=square* ] coordinates { (0.05,0.00) (0.1,0.00) (0.3,0.00) (0.5,0.00) (0.8,0.02) }; 
\addplot[ blue, thick, dashed, mark=diamond* ] coordinates { (0.05,0.00) (0.1,0.00) (0.3,0.00) (0.5,0.00) (0.8,0.02) }; 
\end{axis}
\begin{axis}[ name=sem, at={(geo.outer north east)}, anchor=outer north west, xshift=1.6cm, width=0.44\linewidth, height=0.22\linewidth, title={Semantic perturbations}, xlabel={$\tau$}, ylabel={CUB PFR (\%) $\downarrow$}, ylabel style={orange}, yticklabel style={orange}, xmin=0.8,xmax=10.2, ymin=0,ymax=100, xtick={1,2,5,10}, ymajorgrids, grid style={dashed,gray!30}, tick align=outside, ] 
\addplot[ orange, thick, mark=o ] coordinates { (1,42.7) (2,58.4) (5,77.7) (10,90.0) }; \addplot[ orange, thick, dashed, mark=triangle* ] coordinates { (1,45.0) (2,60.9) (5,80.1) (10,90.4) }; \end{axis} 
\begin{axis}[ at={(sem.south west)}, anchor=south west, width=0.44\linewidth, height=0.22\linewidth, xmin=0.8,xmax=10.2, ymin=0,ymax=7, ylabel={RIVAL PFR (\%) $\downarrow$}, ylabel style={blue}, yticklabel style={blue}, axis y line*=right, axis x line=none, ] \addplot[ blue, thick, mark=square* ] coordinates { (1,0.00) (2,0.07) (5,0.72) (10,6.22) }; \addplot[ blue, thick, dashed, mark=diamond* ] coordinates { (1,0.00) (2,0.08) (5,0.73) (10,6.13) }; \end{axis} 
\node[ draw, font=\scriptsize, anchor=north, inner sep=3pt ] at ($(geo.south)!0.5!(sem.south)+(0,-1.2cm)$) { \begin{tabular}{@{}l@{}} \textbf{Color code: } \tikz\draw[orange,fill=orange] (0,0) circle (2pt);\, CUB-200-2011 \quad \tikz\draw[blue,fill=blue] (0,0) circle (2pt);\, RIVAL-10 \\[2pt] \textbf{Line code: } \tikz\draw[black,thick] (0,0)--(0.6,0);\, $f_{\mathrm{std}}$ \quad \tikz\draw[black,thick,dashed] (0,0)--(0.6,0);\, $f_{\mathrm{cbm}}$ 
\end{tabular} }; 
\end{tikzpicture}
\caption{ \textbf{Empirical robustness under geometric and semantic perturbations.} Orange (left axis): CUB-200-2011; blue (right axis): RIVAL-10. Dataset-specific y-axis scales are used for clarity. } 
\label{fig:pfr_geo_sem} 
\end{figure}
\begin{figure}[t]
\centering
\begin{tikzpicture}

\begin{groupplot}[
  group style={
    group size=2 by 1,
    horizontal sep=1.6cm,
  },
  width=0.46\linewidth,
  height=0.22\linewidth,
  ylabel={PFR \% $\downarrow$},
  ymin=20, ymax=100,
  grid=major,
  grid style={dashed,gray!30},
  tick align=outside,
  tick style={black},
]

\nextgroupplot[
  xlabel={Decision Function Sensitivity (DFS)},
]

\addplot[
  only marks,
  mark=*,
  mark size=2.4pt,
  orange
] coordinates {
  (1.28,31.4)
  (1.59,46.6)
  (3.26,76.6)
  (4.08,87.8)
  (4.84,93.2)
};

\addplot[
  only marks,
  mark=*,
  mark size=2.4pt,
  blue
] coordinates {
  (1.87,45.0)
  (2.57,60.9)
  (3.61,80.1)
  (4.46,90.4)
};

\nextgroupplot[
  xlabel={Concept Flip Rate (CFR)},
]

\addplot[
  only marks,
  mark=*,
  mark size=2.4pt,
  orange
] coordinates {
  (0.40,31.4)
  (0.79,46.6)
  (2.14,76.6)
  (3.15,87.8)
  (4.10,93.2)
};

\addplot[
  only marks,
  mark=*,
  mark size=2.4pt,
  blue
] coordinates {
  (1,45.0)
  (2,60.9)
  (5,80.1)
  (10,90.4)
};

\end{groupplot}

\node[
  draw,
  font=\scriptsize,
  anchor=north,
  inner sep=3pt
] at (current bounding box.south) {
\begin{tabular}{@{}l@{}}
\textbf{Color code: }
\tikz\draw[orange,fill=orange] (0,0) circle (2pt);\,Geometric \quad
\tikz\draw[blue,fill=blue] (0,0) circle (2pt);\,Semantic
\end{tabular}
};

\end{tikzpicture}

\caption{\textbf{Prediction sensitivity in concept‑level measures on CUB‑200‑2011.}
}
\label{fig:pfr_dfs_cfr_main}
\end{figure}
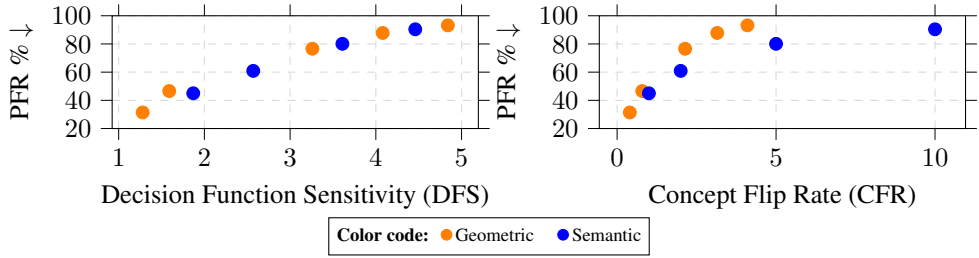

To quantify variability due to subset selection, we repeat the evaluation on $5$ independent random subsets of size 200 drawn uniformly without replacement from the pool (subsets may overlap).
For each subset, both $f_{{std}}$ and $f_{{cbm}}$ are evaluated on the \emph{same} inputs (paired design), yielding five paired measurements per metric.
We then test $H_0:\Delta=0$ using a paired two-sided test across the $5$ subset-level measurements, where $\Delta$ denotes the per-subset difference between methods. When multiple budgets and metrics are tested, we additionally report Holm--Bonferroni corrected $p$-values in Appendix~\ref{app:empirical_stats}.

As shown in \autoref{fig:pfr_geo_sem}, the prediction flip rate (PFR) increases with perturbation strength. On CUB-200-2011, $f_{\mathrm{cbm}}$ exhibits higher PFR than $f_{\mathrm{std}}$ across all evaluated geometric and semantic perturbation budgets. However, after Holm--Bonferroni correction, only the difference at the semantic budget $\tau=5$ remains statistically significant. In contrast, on RIVAL‑10, both models remain nearly perfectly robust over the same perturbation ranges. As shown in \autoref{fig:pfr_dfs_cfr_main}, raw concept flip rates (CFR) alone do not fully account for prediction instability. In contrast, decision function sensitivity (DFS) shows a more consistent relationship with PFR, suggesting that instability in the CBM decision representation better explains prediction flips than the number of concept changes. This arises because CBMs primarily rely on task‑relevant concepts and are relatively insensitive to perturbations of irrelevant ones, aligning with the intuition that concept bottlenecks can act as a form of regularization under structured semantic noise.
Additional analysis of DFS, CFR, and PFR, along with the complete numerical results, is provided in Appendix~\ref{app:dfs_analysis}.

\subsection{Randomised Smoothing Experiments}
\label{sec:rm-exp}

Empirical perturbations characterize sensitivity under sampled inputs but do not provide guarantees
over all perturbations within a neighborhood. To assess whether these trends persist under
worst-case perturbations, we evaluate certified robustness using randomized smoothing
in latent ($\ell_2$) and concept ($\ell_0$) spaces.
Unless otherwise stated, we use
$n=200$ inputs sampled from $\mathcal S_{\mathrm{agree}}$ and estimate smoothed
class probabilities with $N=1000$ Monte Carlo samples. We construct simultaneous one-sided Clopper Pearson confidence bounds with overall confidence level $1-\alpha=95\%$, allocating $\alpha/(2C)$ to each bound, where $C$ is the number of classes. For Bernoulli concept-space certification, the bounds are further tightened using the probability simplex constraint. To quantify variability, we repeat the
evaluation over five independently sampled subsets of size $200$. For latent-space smoothing, we use Gaussian perturbations
$\delta\sim\mathcal N(0,\sigma^2 I)$ with
$\sigma\in\{0.01,0.05,0.10,0.30\}$ and report the average certified
$\ell_2$ radius $R$ over correctly classified, non abstained samples. For
concept space smoothing, each concept is independently flipped with probability
$\rho$, and we report the average certified $\ell_0$ concept flip radius
$\Delta$ over correctly classified, non-abstained samples. In both settings,
smoothed accuracy (SmAcc) counts abstentions as incorrect.

\begin{figure}[t]
\centering
\begin{tikzpicture}

\begin{groupplot}[
  group style={
    group size=4 by 1,
    horizontal sep=1.7cm,
  },
  width=0.25\linewidth,
  height=0.19\linewidth,
  tick align=outside,
  tick style={black},
  grid=major,
  grid style={dashed,gray!30},
]

\nextgroupplot[
  title={(a) Latent},
  xlabel={$\sigma$},
  ylabel={SmAcc (\%) $\uparrow$},
  xmode=log,
  xmin=0.008, xmax=0.4,
  xtick={0.01,0.05,0.10,0.30},
  xticklabels={0.01,0.05,0.10,0.30},
  xticklabel style={font=\scriptsize, rotate=30, anchor=east},
  ymin=55, ymax=102,
  xlabel shift=-10pt,
]

\addplot[orange, thick] coordinates {
  (0.01,96.20) (0.05,88.40) (0.10,81.70) (0.30,62.40)
};
\addplot[orange, thick, dashed] coordinates {
  (0.01,97.40) (0.05,84.90) (0.10,77.20) (0.30,57.10)
};

\addplot[blue, thick] coordinates {
  (0.01,100.0) (0.05,100.0) (0.10,100.0) (0.30,100.0)
};
\addplot[blue, thick, dashed] coordinates {
  (0.01,100.0) (0.05,100.0) (0.10,100.0) (0.30,100.0)
};

\nextgroupplot[
  title={(b) Latent},
  xlabel={$\sigma$},
  ylabel={$R$ $\uparrow$},
  xmode=log,
  xmin=0.008, xmax=0.4,
  xtick={0.01,0.05,0.10,0.30},
  xticklabels={0.01,0.05,0.10,0.30},
  xticklabel style={font=\scriptsize, rotate=30, anchor=east},
  ymin=0, ymax=0.8,
  xlabel shift=-10pt,
]

\addplot[orange, thick] coordinates {
  (0.01,0.0228) (0.05,0.0747) (0.10,0.1034) (0.30,0.1263)
};
\addplot[orange, thick, dashed] coordinates {
  (0.01,0.0215) (0.05,0.0672) (0.10,0.0918) (0.30,0.1183)
};

\addplot[blue, thick] coordinates {
  (0.01,0.0246) (0.05,0.1232) (0.10,0.2463) (0.30,0.7376)
};
\addplot[blue, thick, dashed] coordinates {
  (0.01,0.0246) (0.05,0.1232) (0.10,0.2463) (0.30,0.7375)
};

\nextgroupplot[
  title={(c) Concept},
  xlabel={$\rho$},
  ylabel={SmAcc (\%) $\uparrow$},
  xtick={0.05,0.10,0.20},
  ymin=60, ymax=102,
  xticklabel style={font=\scriptsize},
  xticklabel style={rotate=30, anchor=east},
  xlabel shift=-10pt,
]

\addplot[orange, thick] coordinates {
  (0.05,100.00) (0.10,95.10) (0.20,64.50)
};
\addplot[orange, thick, dashed] coordinates {
  (0.05,100.00) (0.10,95.80) (0.20,63.40)
};

\addplot[blue, thick] coordinates {
  (0.05,100.00) (0.10,100.00) (0.20,100.00)
};
\addplot[blue, thick, dashed] coordinates {
  (0.05,100.00) (0.10,100.00) (0.20,100.00)
};

\nextgroupplot[
  title={(d) Concept},
  xlabel={$\rho$},
  ylabel={$\Delta^\star$ $\uparrow$},
  xtick={0.05,0.10,0.20},
  ymin=0, ymax=3.4,
  xticklabel style={font=\scriptsize},
  xticklabel style={rotate=30, anchor=east},
  xlabel shift=-10pt,
  name=conceptradius,
]

\addplot[orange, thick] coordinates {
  (0.05,0.002) (0.10,0.002) (0.20,0.009)
};
\addplot[orange, thick, dashed] coordinates {
  (0.05,0.001) (0.10,0.001) (0.20,0.002)
};

\addplot[blue, thick] coordinates {
  (0.05,1.000) (0.10,1.834) (0.20,3.000)
};
\addplot[blue, thick, dashed] coordinates {
  (0.05,1.000) (0.10,1.822) (0.20,3.084)
};

\end{groupplot}

\node[
  draw,
  font=\scriptsize,
  anchor=north,
  inner sep=3pt
] at (current bounding box.south) {
\begin{tabular}{@{}l@{}}
\textbf{Color code: }
\tikz\draw[orange,fill=orange] (0,0) circle (2pt);\,CUB-200-2011 \quad
\tikz\draw[blue,fill=blue] (0,0) circle (2pt);\,RIVAL-10 \quad
\textbf{Line code: }
\tikz\draw[black,thick] (0,0)--(0.6,0);\, $f_{\mathrm{std}}$ \quad
\tikz\draw[black,thick,dashed] (0,0)--(0.6,0);\, $f_{\mathrm{cbm}}$
\end{tabular}
};

\end{tikzpicture}
\caption{
\textbf{Certified robustness under latent‑space and concept‑space randomized smoothing.}
}
\label{fig:rm_four_panel}
\end{figure}
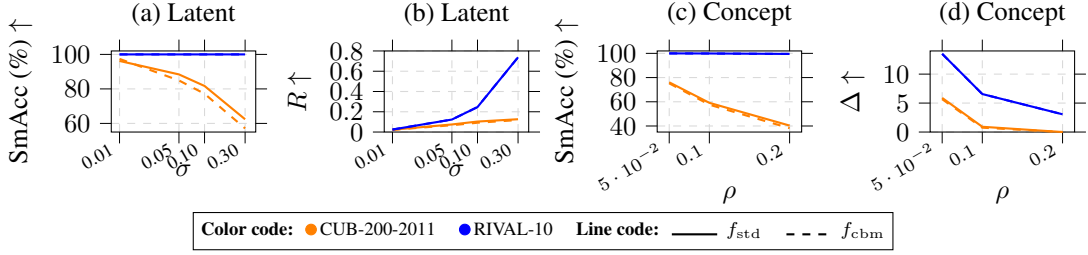

\autoref{fig:rm_four_panel} summarizes the results. Under latent-space smoothing,
the \textit{$f_{std}$ generally achieves higher SmAcc and larger certified radii
than the $f_{cbm}$} on CUB-200-2011, especially as $\sigma$ increases. This indicates
that, on generator native inputs, $f_{cbm}$ does not improve
certifiable robustness to continuous latent perturbations. On the original
RIVAL-10 split, both models are saturated: SmAcc remains at 100\% and certified
radii are nearly identical across all $\sigma$, reflecting the strong class
separability of this benchmark and
indicating that samples remain far from decision boundaries in latent space.

Under concept-space smoothing, increasing the flip probability ($\rho$) can reduce SmAcc while increasing the certified ($\ell_0$) radius, highlighting that \textit{smoothed prediction accuracy and certified robustness capture distinct aspects of model stability}. On CUB-200-2011, the $f_{cbm}$ does not consistently improve certified
concept robustness: although it slightly improves SmAcc for some noise levels,
its certified radius is comparable to or smaller than that of the ResNet
baseline. On RIVAL-10, the certificates are again saturated, motivating the
harder RIVAL-10 variants studied in \autoref{sec:taskC}.

\subsection{Adversarial Evaluation}

We apply the APGD‑CE attack~\citep{croce2020reliable} to the entire pipeline, including the generator, for a practical stress test of adversarial robustness rather than a head‑to‑head robustness comparison. 
As shown in \autoref{fig:attack_summary}, randomized smoothing consistently improves robust accuracy under both latent‑space and concept‑space attacks, with larger gains at higher attack strengths. 
While \autoref{sec:rm-exp} indicates that $f_{cbm}$ does not outperform $f_{std}$ under randomized smoothing in either latent or concept space, adversarial evaluations reveal a complementary pattern: $f_{cbm}$ is more robust to latent‑space attacks but less robust to concept‑space attacks. This reflects the dual role of concept bottlenecks as both a new attack interface and a regularizing mechanism against geometric perturbations, helping resolve prior conflicting conclusions about whether concept bottlenecks introduce robustness.

\def\shiftA{-12pt}
\def\shiftB{-4pt}
\def\shiftC{4pt}
\def\shiftD{12pt}

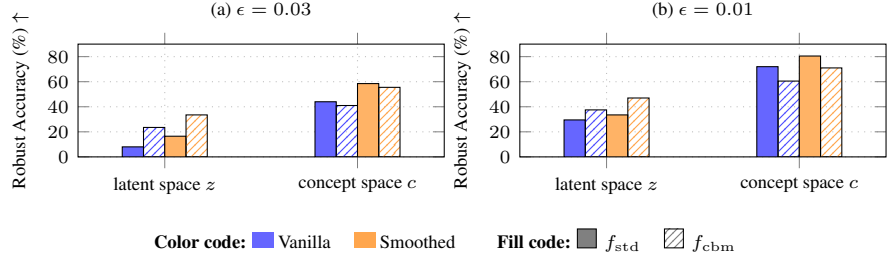
\begin{figure}[t]
\centering
\begin{tikzpicture}

\begin{groupplot}[
  group style={group size=2 by 1, horizontal sep=1.0cm},
  width=0.46\linewidth,
  height=0.22\linewidth,
  ybar,
  ymin=0, ymax=90,
  ylabel={Robust Accuracy (\%) $\uparrow$},
  xtick={1,2},
  xticklabels={latent space $z$, concept space $c$},
  enlarge x limits=0.45,  
  grid=major,
  grid style=dotted,
  tick label style={font=\scriptsize},
  label style={font=\scriptsize},
  title style={font=\scriptsize},
]

\nextgroupplot[
  title={(a) $\epsilon = 0.03$},
]

\addplot[ybar, bar width=8pt, bar shift=\shiftA, fill=blue!60]
coordinates {(1,8) (2,44)};
\addplot[ybar, bar width=8pt, bar shift=\shiftB, fill=blue!70, pattern=north east lines, pattern color=blue!70]
coordinates {(1,23.5) (2,41)};

\addplot[ybar, bar width=8pt, bar shift=\shiftC,fill=orange!60
  ]
coordinates {(1,16.5) (2,58.5)};
\addplot[ybar, bar width=8pt, bar shift=\shiftD,
  fill=orange!70, pattern=north east lines, pattern color=orange!70]
coordinates {(1,33.5) (2,55.5)};

\nextgroupplot[
  title={(b) $\epsilon = 0.01$},
]

\addplot[ybar, bar width=8pt, bar shift=\shiftA, fill=blue!60]
coordinates {(1,29.5) (2,72)};
\addplot[ybar, bar width=8pt, bar shift=\shiftB, fill=blue!70, pattern=north east lines, pattern color=blue!70]
coordinates {(1,37.5) (2,60.5)};

\addplot[ybar, bar width=8pt, bar shift=\shiftC, fill=orange!60]
coordinates {(1,33.5) (2,80.5)};

\addplot[ybar, bar width=8pt, bar shift=\shiftD,
  fill=orange!70, pattern=north east lines, pattern color=orange!70]
coordinates {(1,47) (2,71)};

\end{groupplot}

\node[
  font=\scriptsize,
  anchor=north,
  yshift=-6pt
] at (current bounding box.south) {
\textbf{Color code:}
\textcolor{blue!60}{\rule{8pt}{6pt}} Vanilla \quad
\textcolor{orange!70}{\rule{8pt}{6pt}}  Smoothed
\qquad
\textbf{Fill code:}
\tikz\draw[fill=gray, draw=black] (0,0) rectangle (0.25,0.25);\, $f_{\mathrm{std}}$ \quad
\tikz\draw[
  fill=gray,
  pattern=north east lines,
  pattern color=gray,
  draw=black
] (0,0) rectangle (0.25,0.25);\, $f_{\mathrm{cbm}}$
};

\end{tikzpicture}

\caption{\textbf{Adversarial robustness under APGD‑CE~\citep{croce2020reliable} attacks on the CUB generated dataset.}
}
\label{fig:attack_summary}
\end{figure}

\takeaway{\textbf{Takeaway.} (1) Concept bottlenecks do not uniformly improve prediction-level robustness. Although they can exhibit greater stability under certain semantic perturbation settings, this behavior depends on the task configuration and is not consistently observed.(2) Concept bottlenecks introduce a robustness trade‑off: they improve robustness against geometric adversarial perturbations while increasing vulnerability to concept-space attacks. Our unified framework explains previously contradictory observations and clarifies how future designs can better exploit the robustness trade‑off.}
\section{Robustness under Task Conditions}\label{sec:taskC}

The preceding sections show that concept bottleneck robustness varies substantially across datasets, despite using identical model architectures.
This suggests that robustness is shaped not only by the presence of a bottleneck,
but also by task-level factors.
To isolate these effects, we conduct controlled studies on RIVAL-10 by varying (i) class structure and semantic similarity, and (ii) concept vocabulary size,
while keeping the backbone and training protocol fixed.

\subsection{Effect of Class Structure and Semantic Similarity}

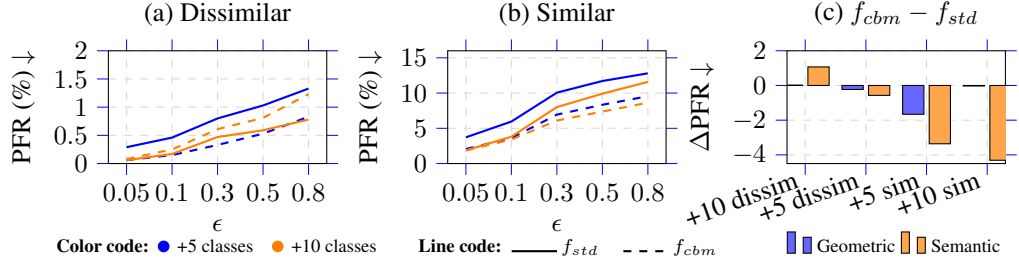
\begin{figure}[t]
\centering
\begin{tikzpicture}

\begin{groupplot}[
  group style={
    group size=3 by 1,
    horizontal sep=1.6cm,
  },
  width=0.32\linewidth,
  height=0.22\linewidth,
  ymin=0, 
  ylabel={PFR (\%) $\downarrow$},
  grid=major,
  grid style={dashed,gray!30},
  tick align=outside,
  tick style={blue},
]

\nextgroupplot[
  title={(a) Dissimilar},
  xlabel={$\epsilon$},
  ymax=2,
  xtick={1,2,3,4,5},
  xticklabels={$0.05$,$0.1$,$0.3$,$0.5$,$0.8$},
  legend to name=shaorangelegend,
]

\addplot[blue, thick] coordinates {
  (1,0.29) (2,0.46) (3,0.80) (4,1.03) (5,1.33)
};
\addlegendentry{$f_{\mathrm{std}}$}

\addplot[blue, thick, dashed] coordinates {
  (1,0.06) (2,0.15) (3,0.33) (4,0.53) (5,0.83)
};
\addlegendentry{$f_{\mathrm{cbm}}$}

\addplot[orange, thick] coordinates {
  (1,0.06) (2,0.17) (3,0.47) (4,0.59) (5,0.78)
};
\addlegendentry{15c}

\addplot[orange, thick, dashed] coordinates {
  (1,0.08) (2,0.25) (3,0.61) (4,0.81) (5,1.23)
};
\addlegendentry{20c}

\nextgroupplot[
  title={(b) Similar},
  xlabel={$\epsilon$},
  ymax=16,
  xtick={1,2,3,4,5},
  xticklabels={$0.05$,$0.1$,$0.3$,$0.5$,$0.8$},
]

\addplot[blue, thick] coordinates {
  (1,3.72) (2,5.96) (3,10.06) (4,11.71) (5,12.80)
};
\addplot[blue, thick, dashed] coordinates {
  (1,2.06) (2,3.61) (3,6.94) (4,8.34) (5,9.55)
};
\addplot[orange, thick] coordinates {
  (1,1.86) (2,3.81) (3,8.01) (4,9.91) (5,11.59)
};
\addplot[orange, thick, dashed] coordinates {
  (1,1.85) (2,3.40) (3,6.10) (4,7.37) (5,8.63)
};

\nextgroupplot[
  title={(c) $f_{cbm} - f_{std}$},
  ylabel={$\Delta$PFR $\downarrow$},
  ybar,
  bar width=8pt,
  ymin=-4.5, ymax=2,
  xtick={1,2,3,4},
  xticklabels={ +10 dissim, +5 dissim, +5 sim , +10 sim},
  xticklabel style={rotate=20, anchor=east},
  legend style={
    at={(0.5,-0.55)},
    anchor=north,
    legend columns=2,
    draw=none,
    font=\scriptsize
  },
]

\addplot[fill=blue!60] coordinates {
  (1,0.02) (2,-0.23) (3,-1.66) (4,-0.03)
};
\addlegendentry{Geometric}

\addplot[fill=orange!70] coordinates {
  (1,1.07) (2,-0.57) (3,-3.36) (4,-4.31)
};
\addlegendentry{Semantic}

\end{groupplot}


\node[
  font=\scriptsize,
  anchor=north,
  yshift=-25pt
] at ($(group c1r1.south)!0.5!(group c2r1.south)$) {
\textbf{Color code:}\,
\tikz\draw[blue,fill=blue] (0,0) circle (2pt);\, +5 classes\quad
\tikz\draw[orange,fill=orange] (0,0) circle (2pt);\, +10 classes
\qquad
\textbf{Line code:}\,
\tikz\draw[black,thick] (0,0)--(0.6,0);\,$f_{std}$ \quad
\tikz\draw[black,thick,dashed] (0,0)--(0.6,0);\,$f_{cbm}$
};

\end{tikzpicture}

\caption{\textbf{Effect of class structure and semantic similarity on task‑conditional robustness under geometric perturbations.}
(a,b) PFR \emph{v.s.} $\epsilon$ for dissimilar and similar class settings.
(c) Difference in PFR (denoted as $\Delta$ PFR) between $f_{cbm}$ and $f_{std}$ when $\epsilon = 0.8$. Lower  $\Delta$ PFR favors $f_{cbm}$.
}
\label{fig:task_structure_threepanel}
\end{figure}

To study the effect of class similarity on robustness, we augment the original RIVAL‑10 classification task with additional classes designed to vary semantic proximity to the original categories. Specifically, we construct two groups of new classes: similar classes, which are semantically close to the original RIVAL‑10 categories, and dissimilar classes, which are semantically distant. Using these groups, we define four task configurations by adding 5 (or 10) similar (or dissimilar) classes, to the original 10‑class task, yielding classification problems of increasing size and semantic difficulty. Across all configurations, we retain the original 18 concept attributes provided by RIVAL‑10, ensuring that only the class structure changes while the concept vocabulary remains fixed. We then evaluate how robustness varies across these controlled settings.

Across all settings, class semantic similarity emerges as the dominant factor shaping robustness differences between $f_{std}$ and $f_{cbm}$. 
\autoref{fig:task_structure_threepanel} shows that semantic class dissimilarity generally improves robustness, with classifiers exhibiting lower prediction instability when classes are well separated. Across configurations, $f_{cbm}$ tends to match or slightly outperform $f_{std}$ under geometric perturbations and mild semantic changes, but becomes increasingly sensitive under large semantic perturbations, reflecting greater vulnerability at the concept level. 
The certified robustness results provide a complementary perspective to the empirical robustness evaluations, with differences between the two models depending on the perturbation space and task configuration. In addition, randomized smoothing yields larger robustness gains for $f_{cbm}$ in several settings, suggesting that smoothing can interact favorably with structured concept representations, albeit without providing a uniform robustness advantage.

\subsection{Effect of Concept Vocabulary Size}

We study the effect of concept set size on robustness using the original RIVAL‑10 task. Starting from the original set of 18 annotated concepts, we use a large language model to generate an additional 18 concepts that are semantically relevant to these classes. This yields three concept configurations with increasing granularity: 18, 27, and 36 concepts. For each configuration, we evaluate robustness while keeping the classification task fixed, allowing us to isolate the impact of concept vocabulary size on model robustness.

\begin{figure*}[t]
\centering
\begin{tikzpicture}

\node (A) {
\begin{subfigure}{0.48\linewidth}
\centering
\begin{tikzpicture}
\begin{axis}[
    title={(a) Geometric perturbation},
    width=\linewidth,
    height=2.6cm,
    xlabel={$\epsilon$},
    ylabel={ PFR \% $\downarrow$},
    ymin=0, ymax=0.065,
    symbolic x coords={0.05,0.10,0.30,0.50,0.80},
    xtick=data,
    xticklabel style={font=\scriptsize},
    grid=major,
    grid style=dotted,
    tick label style={font=\scriptsize},
    label style={font=\scriptsize},
    title style={font=\scriptsize},
    legend to name=shaorangelegend,
    legend columns=2,
    legend style={font=\scriptsize},
]

\addplot[blue, thick] coordinates {
 (0.05,0.00) (0.10,0.01) (0.30,0.01) (0.50,0.01) (0.80,0.04)
};
\addlegendentry{$f_{\mathrm{std}}$}

\addplot[blue, thick, dashed] coordinates {
 (0.05,0.00) (0.10,0.00) (0.30,0.00) (0.50,0.01) (0.80,0.04)
};
\addlegendentry{$f_{\mathrm{cbm}}$}

\addplot[orange, thick] coordinates {
 (0.05,0.00) (0.10,0.00) (0.30,0.00) (0.50,0.01) (0.80,0.04)
};
\addlegendentry{27c}

\addplot[orange, thick, dashed] coordinates {
 (0.05,0.00) (0.10,0.00) (0.30,0.00) (0.50,0.00) (0.80,0.06)
};
\addlegendentry{36c}

\end{axis}
\end{tikzpicture}
\end{subfigure}
};

\node (B) [right=0.4cm of A] {
\begin{subfigure}{0.48\linewidth}
\centering
\begin{tikzpicture}
\begin{axis}[
    title={(b) Semantic perturbation},
    width=\linewidth,
    height=2.6cm,
    xlabel={$\tau$},
    ylabel={PFR \% $\downarrow$},
    ymin=0, ymax=10,
    xtick={1,2,5,10},
    grid=major,
    grid style=dotted,
    tick label style={font=\scriptsize},
    label style={font=\scriptsize},
    title style={font=\scriptsize},
]

\addplot[blue, thick] coordinates {(1,0.03)(2,0.16)(5,1.54)(10,9.29)};
\addplot[blue, thick, dashed] coordinates {(1,0.02)(2,0.15)(5,1.53)(10,9.42)};
\addplot[orange, thick] coordinates {(1,0.02)(2,0.02)(5,0.30)(10,3.11)};
\addplot[orange, thick, dashed] coordinates {(1,0.00)(2,0.02)(5,0.37)(10,2.65)};

\end{axis}
\end{tikzpicture}
\end{subfigure}
};

\node (C) [below=-0.2cm of A] {
\begin{subfigure}{0.48\linewidth}
\centering
\begin{tikzpicture}
\begin{axis}[
    title={(c) Certified in Latent space},
    width=\linewidth,
    height=2.6cm,
    xlabel={$\sigma$},
    ylabel={$R$ $\uparrow$},
    ymin=0, ymax=0.85,
    symbolic x coords={0.01,0.05,0.10,0.30},
    xtick=data,
    xticklabel style={font=\scriptsize},
    grid=major,
    grid style=dotted,
    tick label style={font=\scriptsize},
    label style={font=\scriptsize},
    title style={font=\scriptsize},
]

\addplot[blue, thick] coordinates {
    (0.01,0.0251)
    (0.05,0.1257)
    (0.10,0.2513)
    (0.30,0.7511)
};

\addplot[blue, thick, dashed] coordinates {
    (0.01,0.0251)
    (0.05,0.1256)
    (0.10,0.2511)
    (0.30,0.7498)
};

\addplot[orange, thick] coordinates {
    (0.01,0.0251)
    (0.05,0.1256)
    (0.10,0.2511)
    (0.30,0.7518)
};

\addplot[orange, thick, dashed] coordinates {
    (0.01,0.0251)
    (0.05,0.1256)
    (0.10,0.2512)
    (0.30,0.7513)
};

\end{axis}
\end{tikzpicture}
\end{subfigure}
};

\node (D) [right=0.4cm of C] {
\begin{subfigure}{0.48\linewidth}
\centering
\begin{tikzpicture}
\begin{axis}[
    title={(d) Certified in Concept space},
    width=\linewidth,
    height=2.6cm,
    xlabel={$\rho$},
    ylabel={$\Delta^\star$ $\uparrow$},
    ymin=0, ymax=2.4,
    ybar,
    bar width=5pt,
    symbolic x coords={0.05,0.10,0.20},
    xtick=data,
    grid=major,
    grid style=dotted,
    tick label style={font=\scriptsize},
    label style={font=\scriptsize},
    title style={font=\scriptsize},
]

\addplot[fill=blue] coordinates {
    (0.05,0.983)
    (0.10,1.449)
    (0.20,2.002)
};

\addplot[fill=blue, pattern=north east lines] coordinates {
    (0.05,0.981)
    (0.10,1.451)
    (0.20,2.011)
};

\addplot[fill=orange] coordinates {
    (0.05,0.996)
    (0.10,1.657)
    (0.20,2.152)
};

\addplot[
    fill=orange,
    pattern=north east lines,
    pattern color=orange
] coordinates {
    (0.05,0.996)
    (0.10,1.615)
    (0.20,2.200)
};

\end{axis}
\end{tikzpicture}
\end{subfigure}
};


\node[
  font=\scriptsize,
  anchor=north west,
  yshift=6pt
] at ($(C.south west)+(0,-0.1)$) {
\textbf{Color code:}\,
\tikz\draw[blue,fill=blue] (0,0) circle (2pt);\, K=27 \quad
\tikz\draw[orange,fill=orange] (0,0) circle (2pt);\, K=36
\qquad
\textbf{Line code:}\,
\tikz\draw[black,thick] (0,0)--(0.5,0);\, $f_{std}$\quad
\tikz\draw[black,thick,dashed] (0,0)--(0.5,0);\,$f_{cbm}$
};

\node[
  font=\scriptsize,
  anchor=north east,
  yshift=6pt
] at ($(D.south east)+(0,0)$) {
\textbf{Color:}\,
\tikz\draw[fill=blue] (0,0) rectangle (0.2,0.15);\,  K=27\quad
\tikz\draw[fill=orange] (0,0) rectangle (0.2,0.15);\,  K=36\quad
\textbf{Fill:}
\tikz\draw[fill=gray, draw=black] (0,0) rectangle (0.25,0.25);\, $f_{\mathrm{std}}$ \quad
\tikz\draw[
  fill=gray,
  pattern=north east lines,
  pattern color=gray,
  draw=black
] (0,0) rectangle (0.25,0.25);\, $f_{\mathrm{cbm}}$
};

\end{tikzpicture}

\caption{
\textbf{Impact of concept vocabulary size on robustness.}
Concept-vocabulary ablation on RIVAL-10.
(a,b) Empirical PFR under geometric latent perturbations and semantic concept flips.
(c,d) Certified robustness under Gaussian latent smoothing and Bernoulli concept smoothing.
}
\label{fig:concept_size_radius}
\end{figure*}
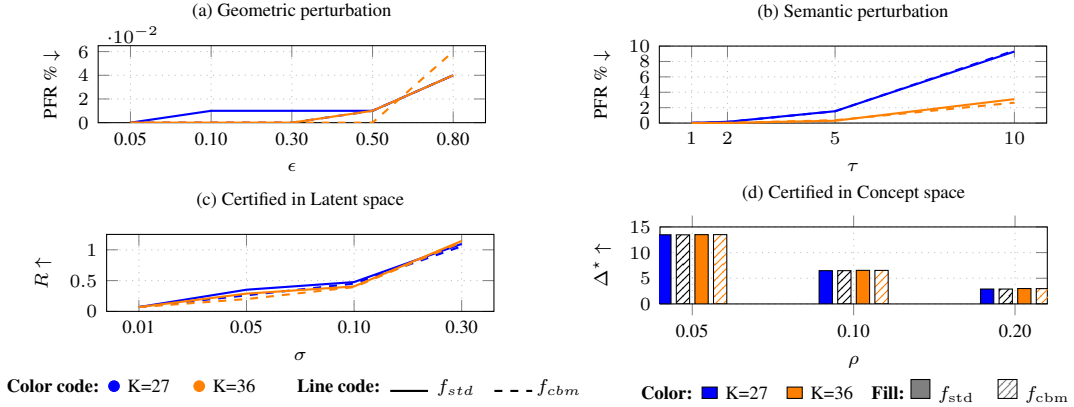

\autoref{fig:concept_size_radius} shows that changing the concept vocabulary affects the robustness behavior of both $f_{\mathrm{std}}$ and $f_{\mathrm{cbm}}$, but the direction of the effect depends on the perturbation setting. Under geometric perturbations, the original RIVAL-10 setting is nearly saturated. Thus, geometric perturbations provide little evidence that either model is more robust. Semantic concept flips are more diagnostic. With 27 concepts, $f_{cbm}$ is slightly more stable at the largest flip budget ($9.29$ vs. $9.42$ PFR), whereas with 36 concepts the trend reverses ($3.11$ vs. $2.65$ PFR). This suggests that the effect of the concept vocabulary is not monotonic. One possible explanation is that, in the 27 concept setting, some flipped concepts are redundant or weakly used by the $f_{cbm}$ decision head, allowing the model to absorb large semantic flips slightly better. In the 36 concept setting, however, the additional concepts may introduce more correlated or unevenly useful attributes; perturbing them can destabilize the $f_{cbm}$ concept representation without yielding a consistent semantic robustness benefit. The randomized smoothing results further show that certified robustness depends on the perturbation regime and concept configuration, without establishing a consistent advantage for either model. Under latent Gaussian smoothing, both classifiers achieve nearly identical certified radii across the 27- and 36-concept configurations, with their curves largely overlapping. This suggests that, in these highly stable settings, latent-space certification provides limited discrimination between model architectures or concept vocabulary sizes. In contrast, concept-space smoothing reveals more noticeable differences across concept configurations. The 36-concept setting generally achieves larger average certified Hamming radii than the 27-concept setting, while the relative differences between $f_{\mathrm{std}}$ and $f_{\mathrm{cbm}}$ remain small and vary with the smoothing probability. These observations indicate that a larger concept vocabulary can influence certified semantic robustness without consistently benefiting the CBM relative to the standard classifier. Overall, concept vocabulary size changes the measured robustness profile, but does not induce a uniform robustness advantage for $f_{\mathrm{cbm}}$.

\takeaway{\textbf{Takeaway.} (1) Lower inter-class similarity is generally associated with higher classifier robustness. (2) Under higher class similarity, concept bottlenecks can retain greater prediction stability or smoothed accuracy in certain settings, although this does not consistently translate into larger certified radii. (3) Increasing the number of concepts does not monotonically improve robustness: additional concepts can help when perturbations target redundant attributes, but may also introduce instability when concepts are correlated or weakly grounded.}

\paragraph{Design Implications for Robust Concept Bottlenecks.}
These results indicate that robustness differences between $f_{std}$ and $f_{cbm}$ are primarily governed by task structure. This suggests that concept bottlenecks are particularly beneficial for classification tasks with high inter‑class similarity, provided that the concept vocabulary is chosen carefully. Our findings further highlight that robustness should be treated as a key criterion when selecting and designing concept sets. While concept bottlenecks may reduce prediction sensitivity under certain semantic perturbations, they can also become more vulnerable under larger concept interventions or adversarial semantic attacks. Consequently, improving concept bottleneck robustness requires careful consideration of task structure, concept vocabulary design, and perturbation specific vulnerabilities.

\section{Limitation and Conclusion}

\paragraph{Limitation.} Our analysis relies on generator‑native inputs to disentangle geometric and semantic perturbations. While this enables controlled robustness evaluation, results may differ for real‑world inputs outside the generator manifold. We focus on post‑hoc CBMs with frozen backbones; end‑to‑end CBMs or alternative concept learning paradigms may exhibit different behaviors. Finally, our conclusions are limited to robustness under structured geometric and semantic perturbations and do not replace standard pixel‑space adversarial robustness evaluations.

\paragraph{Conclusion.}
We introduced a generator-based framework that disentangles continuous latent/geometric perturbations from discrete concept/semantic interventions and used it to systematically evaluate robustness under empirical, certified, and adversarial criteria. Across CUB-200-2011 and RIVAL-10 variants, we find that concept bottlenecks do not confer uniform robustness improvements. Instead, they reshape where and how sensitivity emerges, with robustness depending strongly on the perturbation space, robustness criterion, task structure, and concept vocabulary design.

Within our controlled evaluation setting, these results clarify when concept bottlenecks suppress irrelevant variation and when they introduce additional sensitivity through the semantic interface. In particular, class semantic similarity and concept granularity emerge as important design factors governing this trade-off. Our findings reinforce that interpretability and robustness are distinct objectives: concept bottlenecks should not be treated as an implicit robustness mechanism, but as a modeling choice whose robustness consequences depend on how concepts are defined and used.

Finally, although CBMs provide our primary analytical setting, the additional PixPNet experiments show that the same generator-defined perturbation-to-prediction evaluation and randomized-smoothing certification pipeline can be applied to a prototype-based interpretable classifier without requiring alignment between heterogeneous internal representations. This broader applicability suggests a path toward evaluating robustness across a wider family of interpretable architectures while preserving the controlled perturbation framework developed here.

\paragraph{Acknowledgment}
This work was supported by The Royal Society Grant (Ensuring Trustworthy AI: Robustness Certification for Large Language Models)[Reference RGS\textbackslash R2\textbackslash 252444]. This work also received support from DFG under grant No.~389792660 (TRR~248), and No.~547583482. GJ is supported by the University of Macau under grants MYRG-SRG2026-00033-FIC.
This work was supported by the National Natural Science Foundation of
China (NSF) (T2422015, 62306212) and the Beijing-Tianjin-Hebei Natural Science Foundation Cooperation Project (25JJJJC0009).


\bibliography{ref}
\bibliographystyle{plainnat}


\newpage


\appendix
\newpage

\onecolumn

To support reproducibility and completeness, the supplementary material
provides extended background and related work (\autoref{app:related}),
additional methodological details (\autoref{app:method}), including full
randomized smoothing proofs and algorithms, comprehensive experimental
results (\autoref{app:experiments}), and cross-family applicability experiments
(\autoref{app:cross_family}).

\section{Background and Related Work}
\label{app:related}
\paragraph{Concept Bottleneck Models.}
Concept Bottleneck Models (CBMs) were introduced by \citet{koh2020concept} as a framework that decomposes prediction into concept extraction followed by concept‑based decision making. Building on this idea, subsequent work has explored CBMs along several directions. Post‑hoc CBMs (PCBM) leverage frozen vision or multimodal backbones and add lightweight projection layers to obtain concept representations~\citep{yuksekgonul2022post}. Other approaches focus on improving the quality, uncertainty, or robustness of learned concepts~\citep{vandenhirtz2024stochastic,kim2023probabilistic,havasi2022addressing,zhang2026slcbmenhancingconceptbottleneck}, while additional work aims to increase the flexibility and adaptability of concept representations through interactive or editable mechanisms~\citep{chauhan2023interactive,hu2024editable,shang2024incremental}.
\paragraph{Robustness of Concept Bottlenecks.}
Early work on interpretable models highlighted connections between interpretability and robustness, with Self‑Explaining Neural Networks (SENN) enforcing robustness as the stability of explanations under small perturbations~\citep{sawada2022c}. More recent studies directly examine robustness in Concept Bottleneck Models (CBMs). \citet{sinha2023understanding} show that interpretability alone does not guarantee robustness, demonstrating that concept representations can be fragile under adversarial perturbations and that explicit concept interfaces introduce new attack surfaces, revealing a tension between concept‑level stability and adversarial robustness. In contrast, \citet{rasheed2024exploring} find that CBMs, particularly sequentially trained variants, can achieve higher adversarial accuracy than end‑to‑end models under weak to moderate attacks, suggesting that conceptual bottlenecks may act as an information‑filtering regularizer at the prediction level. Related observations beyond CBMs further indicate that highly interpretable representations can remain non‑robust~\citep{li2026evaluating}, while recent extensions such as language‑guided and flexible CBMs improve scalability and adaptability without addressing robustness explicitly~\citep{yu2025language,du2026flexible}. Overall, prior work reports seemingly conflicting conclusions, which we argue arise from differences in robustness definitions, perturbation models, and task settings rather than fundamental inconsistencies.
\paragraph{Concept-Based Generative Models.}
\label{sec:cbae}
Prior work explores disentangled or concept‑based generative models for controllable generation~\citep{chen2018isolating,ding2020guided,higgins2017beta,ismail2023concept}, while CB‑AE~\citep{kulkarni2025interpretable} enables concept‑level control by augmenting frozen pretrained generators with minimal supervision. In this work, we use CB‑AE~\citep{kulkarni2025interpretable} solely as an experimental instrument to enable controlled geometric and semantic perturbations.

\paragraph{Robustness Beyond Pixel Space.}

Deep neural networks are known to be highly sensitive to small input perturbations, motivating extensive work on adversarial robustness and robustness evaluation~\citep{szegedy,carlini2019evaluating,meng2022adversarial,huang2020survey}. However, most studies focus on pixel‑level perturbations, which often lack semantic meaning and do not reflect changes in underlying factors of variation. Recent work has therefore explored robustness beyond the pixel space, including robustness to perturbations in learned representations or structured transformations~\citep{hao2022gsmooth,wang2021certified,wagh2022evaluating,jin2022enhancing,jin2025enhancing}. Despite this progress, how architectural choices such as concept‑aligned or interpretable representations affect robustness under semantic perturbations remains underexplored, motivating our focus on concept‑level robustness.

\paragraph{Randomized Smoothing for Certified Robustness.}

Randomized smoothing \citep{cohen2019certified} was developed to evaluate probabilistic certified robustness for classification tasks. By constructing a smoothed classifier $\hat{f}(x)$ through aggregation of predictions over randomized perturbations, it can produce the most probable prediction of the base classifier $f(x)$ over perturbed inputs from Gaussian noise in a test instance.  The smoothed classifier $\hat{f}(x)$ is supposed to be provably robust to $l_2$-norm bounded perturbations within a certain radius $R$:
\begin{theorem}\label{the:cohen}
 \citep{cohen2019certified} For a classifier $f: \mathbb{R} \to \mathcal{Y}$, suppose $y \in \mathcal{Y}$, let $\delta \sim \mathcal{N}(0, \sigma^2 I)$. Suppose the top class $A$ is predicted with probability $p_A$, and the runner-up prediceted class $B$, with probability $p_B$. The smoothed classifier be 
 $\hat{f}(\vx):=\mathop{\arg\max}\limits_y \mathbb{P}(f(\vx+\delta)=y)$,
 suppose $\underline{p_A},\overline{p_B} \in [0,1]$, if
 \begin{equation}
     \mathbb{P} (f(\vx+\delta)=y_A)\geq \underline{p_A} \geq \overline{p_B} \geq \mathop{\max}\limits_{y\neq y_A}\mathbb{P}(f(\vx+\delta)=y),
 \end{equation}
 then $\hat{f}(\vx+\epsilon)=y_A$ for all $||\epsilon ||_2 \leq R$, where
 \begin{equation}
     R=\frac{\sigma}{2}(\Phi^{-1}(\underline{p_A})-\Phi^{-1}(\overline{p_B})).
 \end{equation}
\end{theorem}
Here $\Phi^{-1}$ is the inverse cumulative distribution function (CDF) of the normal distribution.

This approach has been applied primarily in pixel space to certify robustness against $\ell_p$-bounded adversarial perturbations, and has been extended to different noise distributions and norm settings. While most work applies smoothing in pixel space, recent studies have explored extensions to learned representations or structured perturbations~\citep{hao2022gsmooth}.
In this work, we use randomized smoothing not as a defense mechanism, but as a comparative evaluation tool to assess certified robustness of concept‑based and standard classifiers under controlled geometric and semantic perturbations.


\section{Methodology}
\label{app:method}

\subsection{Empirical Robustness Characterization}
\label{app:method-exp}
\begin{figure*}
    \centering
    \includegraphics[width=0.85\linewidth, trim=0 3.5cm 0 3.5cm, clip]{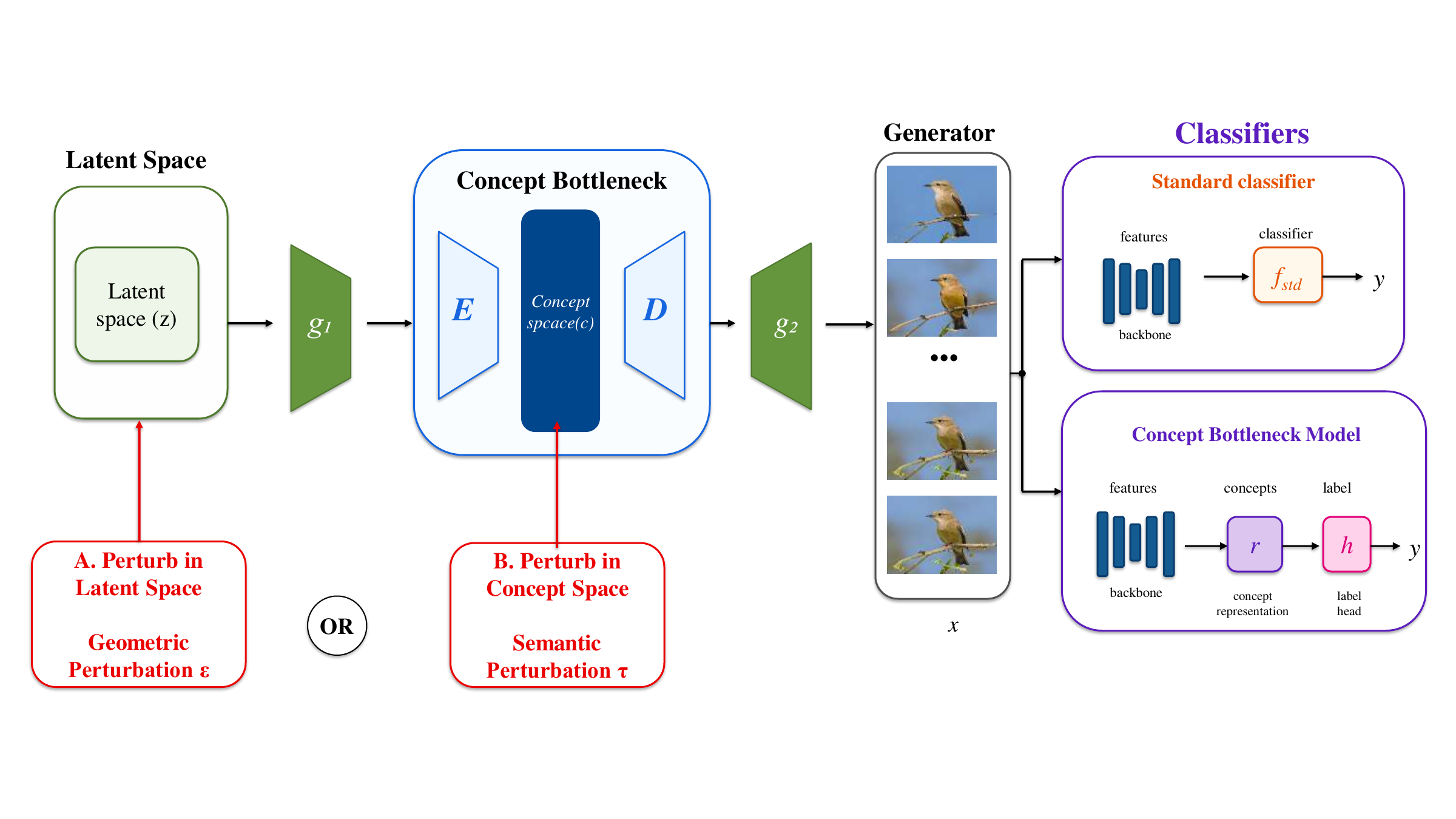}
    \caption{\textbf{Generator-based evaluation framework for geometric vs.\ semantic perturbations.} (i) geometric perturbations by adding latent-space noise within an $\ell_2$ budget $\epsilon$,
and (ii) semantic perturbations by flipping up to $\tau$ binary concepts in concept space before image generation.
}
    \label{fig:framework}
\end{figure*}

As illustrated in \autoref{fig:framework}, we introduce a generator‑based evaluation framework that disentangles continuous latent‑space perturbations from discrete concept‑level interventions. In the latent space, geometric perturbations are applied by modifying the corresponding latent vector $\vz$, while in the concept space, semantic interventions are performed by flipping entries in the concept representation $\vc$. These operations yield controlled sets of generated images reflecting geometric and semantic perturbations, which we use to systematically analyze the robustness of classifiers with and without concept bottlenecks.

\subsection{Latent-Space Gaussian Smoothing}
\label{app:latent_smoothing}

Given a latent representation $\vz$, the latent-smoothed classifier is
\begin{equation}
\hat f_{\mathrm G}(\vz)
=
\arg\max_{y\in\mathcal Y}
\mathbb P_{\delta\sim\mathcal N(0,\sigma^2 I)}
\left[
f(g(\vz+\delta))=y
\right].
\end{equation}
Let $A$ and $B$ denote the top and runner-up classes with probabilities
\begin{equation}
p_A
=
\mathbb P_{\delta}
\left[
f(g(\vz+\delta))=A
\right],
\qquad
p_B
=
\mathbb P_{\delta}
\left[
f(g(\vz+\delta))=B
\right].
\end{equation}
By the Gaussian randomized smoothing certificate of \citet{cohen2019certified}, if $p_A>p_B$, then
$\hat f_{\mathrm G}(\vz+\epsilon)=A$ for all
\begin{equation}
\|\epsilon\|_2
\le
\frac{\sigma}{2}
\left(
\Phi^{-1}(p_A)-\Phi^{-1}(p_B)
\right).
\end{equation}

In practice, $p_A$ and $p_B$ are estimated by Monte Carlo sampling. Given
$N$ i.i.d. samples $\delta_1,\ldots,\delta_N\sim\mathcal N(0,\sigma^2I)$, we
estimate
\begin{equation}
\widehat p_y(\vz)
=
\frac{1}{N}
\sum_{i=1}^N
\mathbb I
\left[
f(g(\vz+\delta_i))=y
\right].
\end{equation}
We construct simultaneous one-sided Clopper--Pearson confidence bounds for all $C$ classes, allocating an error probability of $\alpha/(2C)$ to each bound. The lower bound for the top-voted class and the maximum upper bound among competing classes are then used to compute the certified Gaussian radius. The empirical certified radius is
\begin{equation}
\widehat R_{\mathrm G}
=
\frac{\sigma}{2}
\left(
\Phi^{-1}(\underline p_A)
-
\Phi^{-1}(\overline p_B)
\right).
\end{equation}
The certificate holds with probability at least $1-\alpha$ over the sampling
procedure.

\subsection{Concept-Space Bernoulli Flip Smoothing}
\label{app:concept_smoothing}

Let $\vc\in\mathbb R^K$ denote a concept representation and let
$\mathcal C(\vc)\in\{0,1\}^K$ be its binarized concept vector. For paired concept
coordinates $[c_i^+,c_i^-]$, flipping the $i$-th concept swaps
$[c_i^+,c_i^-]$ into $[c_i^-,c_i^+]$.

We define Bernoulli concept-flip noise by sampling
\begin{equation}
\eta_i\sim\mathrm{Bernoulli}(\rho),
\qquad i=1,\ldots,K,
\end{equation}
independently across concepts. Let $\mathcal F(\vc,\eta)$ denote the concept
representation obtained after flipping all coordinates with $\eta_i=1$.
The concept-smoothed class probabilities are
\begin{equation}
p_y(\vc)
=
\mathbb P_{\eta\sim\mathrm{Bernoulli}(\rho)^K}
\left[
f\!\left(g_2(D(\mathcal F(\vc,\eta)))\right)=y
\right],
\end{equation}
and the concept-smoothed classifier is
\begin{equation}
\hat f_{\mathrm C}(\vc)
=
\arg\max_{y\in\mathcal Y} p_y(\vc).
\end{equation}

We certify robustness to adversarial concept flips satisfying
\begin{equation}
\|\mathcal C(\tilde{\vc})-\mathcal C(\vc)\|_0\le \Delta.
\end{equation}
Because randomized concept flipping modifies semantic values rather than
masking them, sparse-ablation certificates do not directly apply. We instead
use the Neyman--Pearson smoothing framework of \citet{lee2019tight}
on the Hamming cube.

For a candidate adversarial budget $\Delta$, let $\mathcal L_\Delta$ and
$\mathcal U_\Delta$ denote the Neyman--Pearson transfer functions induced by
the Bernoulli smoothing distribution. Intuitively, $\mathcal L_\Delta(p)$ gives
the smallest possible probability of an event after any $\Delta$-bit adversarial
shift, assuming that the event has clean smoothing probability at least $p$.
Similarly, $\mathcal U_\Delta(p)$ gives the largest possible perturbed
probability of an event whose clean smoothing probability is at most $p$.

Let $y^\star=\arg\max_y p_y(\vc)$ be the top class. Suppose conservative
confidence bounds satisfy
\begin{equation}
p_{y^\star}(\vc)\ge \underline p_{y^\star}(\vc),
\qquad
p_y(\vc)\le \overline p_y(\vc),
\qquad \forall y\neq y^\star .
\end{equation}
If
\begin{equation}
\mathcal L_\Delta
\left(
\underline p_{y^\star}(\vc)
\right)
>
\mathcal U_\Delta
\left(
\overline p_y(\vc)
\right),
\qquad
\forall y\neq y^\star,
\end{equation}
then with probability at least $1-\alpha$ over Monte Carlo estimation,
\begin{equation}
\hat f_{\mathrm C}(\tilde{\vc})=y^\star,
\qquad
\forall \tilde{\vc}:
\|\mathcal C(\tilde{\vc})-\mathcal C(\vc)\|_0\le \Delta.
\end{equation}

\paragraph{Certified concept-flip radius.}
The certified adversarial concept-flip radius is
\begin{align}
\Delta^\star(\vc)
=
\max\Big\{
\Delta\in\{0,1,\ldots,K\}:
&
\mathcal L_\Delta
\left(
\underline p_{y^\star}(\vc)
\right)
>
\mathcal U_\Delta
\left(
\overline p_y(\vc)
\right),
\nonumber\\
&
\forall y\in\mathcal Y\setminus\{y^\star\}
\Big\}.
\label{eq:certified_concept_radius_app}
\end{align}

\paragraph{Bernoulli probability transfer.}
Each concept $i$ is represented by paired coordinates
$(c_i^+,c_i^-)$. A binary concept flip exchanges the two
coordinates, i.e.,
$T_i(c_i^+,c_i^-)=(c_i^-,c_i^+)$.
The resulting binary intervention space has $K$ coordinates,
although the continuous paired representation has $2K$
entries. Each Bernoulli variable controls one concept pair,
rather than an individual scalar coordinate.

Consider an adversarial mask $a\in\{0,1\}^K$ with
$d=\|a\|_0$. Since successive swaps satisfy
$\mathcal F(\mathcal F(c,a),\eta)
=\mathcal F(c,a\oplus\eta)$,
the original and perturbed smoothing distributions differ
only on the $d$ intervened concepts.

Let $P$ and $Q$ denote these two distributions. If $m$
of the $d$ coordinates are flipped by the smoothing noise
relative to the original state, the corresponding
probability masses are
\begin{equation}
P_m=\binom{d}{m}\rho^m(1-\rho)^{d-m},
\qquad
Q_m=\binom{d}{m}\rho^{d-m}(1-\rho)^m.
\end{equation}
Their likelihood ratio is
\begin{equation}
\frac{Q_m}{P_m}
=\left(\frac{\rho}{1-\rho}\right)^{d-2m}.
\label{eq:bernoulli_lr_app}
\end{equation}

To compute $L_d(p)$, we order the likelihood-ratio
regions by increasing $Q_m/P_m$ and accumulate
probability mass under $P$ until reaching $p$,
allowing a fractional contribution from the last
region. The corresponding mass under $Q$ gives
$L_d(p)$. The upper transfer function is
\begin{equation}
U_d(p)=1-L_d(1-p).
\label{eq:bernoulli_upper_app}
\end{equation}

\paragraph{Confidence bounds and certified radius.}
For each input, let $n_y$ be the number of predictions
assigned to class $y$ among $N$ independently sampled
Bernoulli perturbations, and let $C=|\mathcal Y|$.
We first compute one-sided Clopper--Pearson bounds
$L_y$ and $U_y$ with per-bound error probability
$\alpha/(2C)$.

For the top-voted class
$A=\arg\max_y n_y$, we then tighten the bounds
using the probability-simplex constraint:
\begin{equation}
\underline p_A=
\max\left\{L_A,\,
1-\sum_{y\ne A}U_y\right\},
\end{equation}
\begin{equation}
\overline p_B=
\min\left\{
\max_{y\ne A}U_y,\,
1-\underline p_A
\right\}.
\end{equation}

The Bonferroni correction ensures simultaneous coverage
with probability at least $1-\alpha$, and the simplex
tightening preserves this coverage. For each candidate
Hamming distance $d\in\{0,\ldots,K\}$, we test
\begin{equation}
L_d(\underline p_A)>U_d(\overline p_B).
\label{eq:bernoulli_test_app}
\end{equation}
The certified radius $\Delta^\star$ is the largest
integer for which the condition holds at every distance
up to $\Delta^\star$. If the condition fails at $d=0$,
the classifier abstains.

By the Neyman--Pearson construction, these transfer
bounds apply to every binary intervention mask of
the corresponding Hamming weight. Hence, whenever
the certification condition holds, the smoothed
prediction remains unchanged under all concept-pair
swaps within the certified Hamming radius.

\subsection{Monte Carlo Estimation and Certification Algorithm}
\label{app:cert_algorithm}

For concept-space smoothing, we estimate the smoothed probabilities using
$N$ i.i.d. Bernoulli flip samples. Abstentions occur when the top-class lower
confidence bound does not exceed the runner-up upper confidence bound. The procedure for estimating the smoothed class probabilities is presented in Algorithm~\autoref{alg:concept_cert}.

\begin{algorithm}[t]
\caption{Concept-space $\ell_0$ certification via Bernoulli concept flips}
\label{alg:concept_cert}
\begin{algorithmic}[1]
\REQUIRE Concept representation $\vc$, base classifier $f$,
generator mapping $g_2(D(\cdot))$, flip probability $\rho$,
samples $N$, confidence level $1-\alpha$.

\FOR{$i=1$ to $N$}
    \STATE Sample $\eta_i\sim\mathrm{Bernoulli}(\rho)^K$.
    \STATE Compute $\vc_i'=\mathcal F(\vc,\eta_i)$.
    \STATE Compute $\hat y_i=f(g_2(D(\vc_i')))$.
\ENDFOR

\STATE Compute class counts
$\mathrm{count}[y]=|\{i:\hat y_i=y\}|$
for all $y\in\mathcal Y$.

\STATE Let $y^\star=\arg\max_y \mathrm{count}[y]$.

\STATE Compute simultaneous Clopper--Pearson bounds
$\underline p_y(\vc)$ and $\overline p_y(\vc)$
using Bonferroni correction.

\IF{$\underline p_{y^\star}(\vc)
\leq \max_{y\neq y^\star}\overline p_y(\vc)$}
    \STATE \textbf{return} \textsc{Abstain}
\ENDIF

\STATE $\Delta^\star(\vc)\gets 0$.

\FOR{$\Delta=1$ to $K$}
    \STATE Compute $L_\Delta(\underline p_{y^\star})$
    and $U_\Delta(\overline p_y)$ for all $y\neq y^\star$.

    \IF{$L_\Delta(\underline p_{y^\star})
    >\max_{y\neq y^\star}U_\Delta(\overline p_y)$}
        \STATE $\Delta^\star(\vc)\gets\Delta$.
    \ELSE
        \STATE \textbf{break}
    \ENDIF
\ENDFOR

\STATE \textbf{return} $(y^\star,\Delta^\star(\vc))$.

\end{algorithmic}
\end{algorithm}


\section{Experiments}
\label{app:experiments}
\subsection{Computation of the robustness-gap heatmap.}\label{app:experiments_fig1}
Each entry in Figure~1(b) is computed from the corresponding representative experimental
setting in the main results. For empirical geometric robustness we use $\epsilon=0.3$; for
empirical semantic robustness we use $\tau=5$; for latent-space randomized smoothing we use
$\sigma=0.10$; and for concept-space randomized smoothing we use $\rho=0.10$.

Since lower PFR indicates greater robustness, empirical gaps are computed as
\[
\mathrm{Gap}_{\mathrm{PFR}}
=
\mathrm{PFR}(f_{\mathrm{std}})
-
\mathrm{PFR}(f_{\mathrm{cbm}}).
\]
Thus, a positive PFR gap means that the CBM has fewer prediction flips.

For certified robustness, larger radii indicate greater robustness. Therefore, latent-space
certified gaps are computed as
\[
\mathrm{Gap}_{R}
=
R(f_{\mathrm{cbm}})-R(f_{\mathrm{std}}),
\]
and concept-space certified gaps are computed as
\[
\mathrm{Gap}_{\Delta^\star}
=
\Delta^\star(f_{\mathrm{cbm}})-\Delta^\star(f_{\mathrm{std}}).
\]
Thus, positive values consistently indicate that the CBM is more robust. PFR gaps are reported
in percentage points, while certified-radius gaps are reported in absolute radius units.
\subsection{More Details on Experiments Setups}

\paragraph{Hareware.} We profiled the GPU memory usage of the two main experimental components under the default setting of batch size 8. All experiments were run on a single NVIDIA RTX 4090 GPU with 24GB memory. For the latent-space perturbation experiment, each epsilon setting involves approximately \(400 \times 1000 = 400{,}000\) latent perturbations, with a peak allocated GPU memory of about 5.3GB and a PyTorch reserved memory of about 6.6GB. For the concept-space flipping experiment, the default setting involves approximately \(4 \times 1000 \times 100 = 1{,}600{,}000\) concept perturbations, with a peak allocated GPU memory of about 5.0GB and a PyTorch reserved memory of about 6.3GB. These measurements indicate that, under the reported experimental scale and batch-size setting, a single RTX 4090 24GB GPU is sufficient for the memory requirements of these experiments. To ensure stable execution, we recommend keeping at least 8GB of free GPU memory, and preferably 10--12GB. Each experiment typically uses a single GPU; multiple experiments are run in parallel via separate CUDA\_VISIBLE\_DEVICES configurations

\subsection{Empirical Robustness Results}
\label{app:empirical_stats}

\begin{table*}[thbp]
  \centering
  \small
  \setlength{\tabcolsep}{2.2pt}
  \begin{tabular}{l r | c c c c | c c c c}
  \toprule
  \multicolumn{2}{c|}{\textbf{Perturbation}}
      & \multicolumn{4}{c|}{\textbf{CUB-200-2011}}
      & \multicolumn{4}{c}{\textbf{RIVAL-10}} \\
  \cmidrule(lr){1-2}\cmidrule(lr){3-6}\cmidrule(lr){7-10}
  Type & Budget
      & $\mathrm{CFR}$ & \multicolumn{2}{c}{\textbf{PFR (\%) $\downarrow$}} & $\mathrm{DFS}\downarrow$
& $\mathrm{CFR}$ & \multicolumn{2}{c}{\textbf{PFR (\%) $\downarrow$}} & $\mathrm{DFS}\downarrow$ \\
  \cmidrule(lr){4-5}\cmidrule(lr){8-9}
      & 
      & 
      & $f_{\mathrm{std}}$ & $f_{\mathrm{cbm}}$
      &
      &
      & $f_{\mathrm{std}}$ & $f_{\mathrm{cbm}}$
      & \\
  \midrule
  \multirow{5}{*}{Geo.}
      & $\epsilon=0.05$ & 0.40$\pm$0.01 & 27.1$\pm$2.1 & 31.4$\pm$1.5 & 1.28$\pm$0.03 &
  0.32 $\pm$0.14 & 0.00 & 0.00 & 0.48$\pm$0.01\\
      & $\epsilon=0.1$  & 0.79$\pm$0.02 & 42.2$\pm$2.5 & 46.6$\pm$1.4 & 1.59$\pm$0.01 &
 0.67 $\pm$0.22 & 0.00 & 0.00 & 0.75$\pm$0.01\\
      & $\epsilon=0.3$  & 2.14$\pm$0.04 & 74.6$\pm$1.6 & 76.6$\pm$0.8 & 3.26$\pm$0.05 &
  2.11 $\pm$0.37 & 0.00 & 0.00 & 1.28$\pm$0.03\\
      & $\epsilon=0.5$  & 3.15$\pm$0.04 & 86.3$\pm$1.0 & 87.8$\pm$1.3 & 4.08$\pm$0.04 &
 4.09$\pm$0.47 & 0.00 & 0.00 & 1.62$\pm$0.02\\
      & $\epsilon=0.8$  & 4.10$\pm$0.04 & 92.6$\pm$0.6 & 93.2$\pm$0.3 & 4.84$\pm$0.03 &
  8.36 $\pm$0.55 & 0.02$\pm$0.01 & 0.02$\pm$0.02 & 1.93$\pm$0.04\\
      \midrule
      \multirow{5}{*}{Sem.}
      & $\tau=1$  & 1  & 42.7$\pm$1.6 & 45.0$\pm$0.8 & 1.87$\pm$0.48 & 1  & 0.00 & 0.00
  & 1.84$\pm$0.02\\
      & $\tau=2$  & 2  & 58.4$\pm$1.3 & 60.9$\pm$0.7 & 2.57$\pm$0.50 & 2  &
  0.07$\pm$0.02 & 0.08$\pm$0.03 & 2.28$\pm$0.03\\
      & $\tau=5$  & 5  & 77.7$\pm$0.9 & 80.1$\pm$0.6 & 3.61$\pm$0.60 & 5  &
  0.72$\pm$0.07 & 0.73$\pm$0.09 & 3.14$\pm$0.05\\
      & $\tau=10$ & 10 & 90.0$\pm$0.6 & 90.4$\pm$0.4 & 4.46$\pm$0.81 & 10 &
  6.22$\pm$0.62 & 6.13$\pm$0.70 & 5.20$\pm$0.11\\
      \bottomrule
      \end{tabular}
      \caption{Robustness comparison under geometric (latent $\ell_2$ noise) and
  semantic (concept flip) perturbations.
   }
      \label{tab:sra_merged}
\end{table*}

\autoref{tab:sra_merged} reports the complete empirical robustness results on both CUB‑200‑2011 and RIVAL‑10. 
Here, CFR reflects the surrogate ground‑truth rate of concept flips induced by the generator. We then measure prediction instability via the classifier’s prediction flip rate. For concept bottleneck models, we additionally evaluate robustness in the intermediate concept space. The results show that, although $f_{cbm}$ is less robust than $f_{std}$at the prediction level, its concept representation is comparatively stable: even when up to 10 concepts are semantically flipped, changes in the concept space remain limited.

\paragraph{Statistical Analysis.}
We test for differences in robustness between $f_{\text{std}}$ and $f_{\text{cbm}}$ using a paired design: for each repeat $k$, both classifiers are evaluated on the same subset, yielding paired measurements.
For each perturbation budget, we define $\Delta_k = \mathrm{PFR}_k(f_{\text{std}}) - \mathrm{PFR}_k(f_{\text{cbm}})$ and test the null hypothesis $H_0:\mathbb{E}[\Delta]=0$ with a two-sided paired test across the $K=5$ repeats.
Since we test multiple budgets, we additionally report Holm--Bonferroni corrected $p$-values.
The per-budget $p$-values and corrected values are reported in \autoref{tab:app-sec42-pvalues}.

\begin{table}[t]
\centering
\small
\setlength{\tabcolsep}{6pt}
\begin{tabular}{c c c c c}
\toprule
Type & Budget & $\overline{\Delta}\pm\sigma_{\Delta}$ (pp) & $p$ & $p_{\mathrm{Holm}}$ \\
\midrule
Geo. & $\epsilon=0.05$ & $-4.25 \pm 2.43$ & 0.017 & 0.098 \\
Geo. & $\epsilon=0.10$ & $-4.39 \pm 2.26$ & 0.012 & 0.086 \\
Geo. & $\epsilon=0.30$ & $-2.00 \pm 1.21$ & 0.021 & 0.098 \\
Geo. & $\epsilon=0.50$ & $-1.12 \pm 0.63$ & 0.016 & 0.098 \\
Geo. & $\epsilon=0.80$ & $-0.51 \pm 0.31$ & 0.022 & 0.098 \\
\midrule
Sem. & $\tau=1$  & $-2.33 \pm 1.49$ & 0.025 & 0.098 \\
Sem. & $\tau=2$  & $-2.83 \pm 1.34$ & 0.009 & 0.073 \\
Sem. & $\tau=5$  & $-2.33 \pm 0.84$ & 0.004 & 0.032 \\
Sem. & $\tau=10$ & $-0.84 \pm 0.49$ & 0.018 & 0.098 \\
\bottomrule
\end{tabular}
\caption{Paired significance tests for PFR differences between $f_{\text{std}}$ and $f_{\text{cbm}}$ across $K=5$ subset repeats (each repeat uses $n=200$ images). $\Delta$ is measured in percentage points (pp). We report raw two-sided paired $t$-test $p$-values and Holm--Bonferroni corrected $p_{\mathrm{Holm}}$ across all 9 budgets in the table.}
\label{tab:app-sec42-pvalues}
\end{table}

Across perturbation budgets, stronger perturbations increase PFR for both classifiers and increase DFS, indicating that the perturbations induce larger changes in the CBM concept representation.
The repeated-subset statistics show that the reported trends are stable across different random selections of evaluation images, and the paired tests quantify when the observed differences in PFR between $f_{\text{std}}$ and $f_{\text{cbm}}$ are statistically significant.

\paragraph{Additional Analysis of CFR, PFR, and DFS}
\label{app:dfs_analysis}

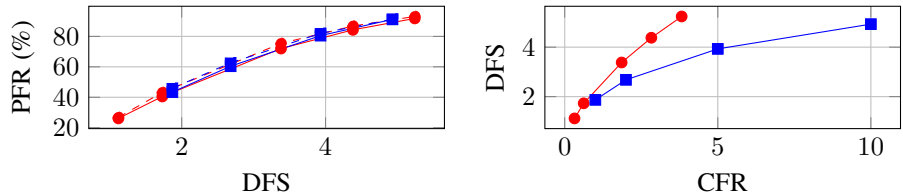
\begin{figure}[thbp]
\centering
\begin{tabular}{c c}

\begin{tikzpicture}
\begin{axis}[
    width=0.45\linewidth,
    height=3.2cm,
    xlabel={DFS},
    ylabel={PFR (\%)},
    grid=major,
    mark size=2pt,
    legend to name=sharedlegend,
    legend columns=2,
]

\addplot[red, solid, mark=*] table[row sep=\\] {
    x y \\
    1.12 26.1 \\
    1.73 40.5 \\
    3.38 72.0 \\
    4.38 84.3 \\
    5.24 91.8 \\
};
\addlegendentry{Geometric + $f$}

\addplot[red, dashed, mark=*] table[row sep=\\] {
    x y \\
    1.12 26.8 \\
    1.73 43.2 \\
    3.38 75.4 \\
    4.38 86.8 \\
    5.24 93.1 \\
};
\addlegendentry{Geometric + $f_{\mathrm{CBM}}$}

\addplot[blue, solid, mark=square*] table[row sep=\\] {
    x y \\
    1.87 43.4 \\
    2.68 60.4 \\
    3.93 80.5 \\
    4.93 91.2 \\
};
\addlegendentry{Semantic + $f$}

\addplot[blue, dashed, mark=square*] table[row sep=\\] {
    x y \\
    1.87 45.9 \\
    2.68 62.6 \\
    3.93 82.0 \\
    4.93 91.3 \\
};
\addlegendentry{Semantic + $f_{\mathrm{CBM}}$}

\end{axis}
\end{tikzpicture}

&
\begin{tikzpicture}
\begin{axis}[
    width=0.45\linewidth,
    height=3.2cm,
    xlabel={CFR},
    ylabel={DFS},
    grid=major,
    mark size=2pt,
]

\addplot[red, solid, mark=*] table[row sep=\\] {
    x y \\
    0.32 1.12 \\
    0.62 1.73 \\
    1.86 3.38 \\
    2.83 4.38 \\
    3.82 5.24 \\
};

\addplot[blue, solid, mark=square*] table[row sep=\\] {
    x y \\
    1   1.87 \\
    2   2.68 \\
    5   3.93 \\
    10  4.93 \\
};

\end{axis}
\end{tikzpicture}

\end{tabular}



\caption{mean PFR \emph{vs} mean DFS (left) and mean DFS \emph{vs} mean CFR (right). {\color{red}Red: Geometric}, {\color{blue}Blue: Semantic}. Solid: $f_{std}$; Dashed: $f_{cbm}$.}
\label{fig:dfs_cfr_pfr}
\end{figure}

As shown in \autoref{fig:dfs_cfr_pfr}, PFR generally increases with CFR, but the relationship is not strictly monotonic.
For example, when the semantic budget is $\tau=5$ (CFR = 5) and the geometric
budget is $\epsilon=0.5$ (CFR = 3.15), the corresponding PFR values are similar,
indicating that higher CFR does not always imply higher prediction instability.

DFS exhibits a more consistent relationship with PFR. Geometric perturbations
tend to induce larger DFS values than semantic perturbations at comparable CFR
levels, suggesting that latent-space perturbations can distort the model decision
representation more strongly than direct concept flips. This may be due to the
nonlinear CB-AE mapping between latent, concept, and image spaces.

\subsection{Randomised Smoothing Experiments}

\begin{table*}[t]
\centering
\small
\setlength{\tabcolsep}{3pt}
\begin{tabular}{l r l | c c | c c}
\toprule
\textbf{Type} & \textbf{Budget} & \textbf{Model}
& \multicolumn{2}{c|}{\textbf{CUB-200-2011}}
& \multicolumn{2}{c}{\textbf{RIVAL-10 (10-class)}} \\
\cmidrule(lr){4-5}\cmidrule(lr){6-7}
& &
& SmAcc (\%) $\uparrow$ & $R/\Delta \uparrow$
& SmAcc (\%) $\uparrow$ & $R/\Delta \uparrow$ \\
\midrule

\multirow{8}{*}{Latent RM}
& \multirow{2}{*}{$\sigma=0.01$}
& $f_{\mathrm{std}}$ & 96.20 $\pm$ 0.84 & \textbf{0.0228 $\pm$ 0.0005} & 100.00 $\pm$ 0.00 & 0.0246 $\pm$ 0.0000 \\
& & $f_{\mathrm{cbm}}$ & \textbf{97.40 $\pm$ 0.55} & 0.0215 $\pm$ 0.0004 & 100.00 $\pm$ 0.00 & 0.0246 $\pm$ 0.0000 \\
\cmidrule(lr){2-7}

& \multirow{2}{*}{$\sigma=0.05$}
& $f_{\mathrm{std}}$ & \textbf{88.40 $\pm$ 1.39} & \textbf{0.0747 $\pm$ 0.0031} & 100.00 $\pm$ 0.00 & 0.1232 $\pm$ 0.0000 \\
& & $f_{\mathrm{cbm}}$ & 84.90 $\pm$ 1.67 & 0.0672 $\pm$ 0.0026 & 100.00 $\pm$ 0.00 & 0.1232 $\pm$ 0.0000 \\
\cmidrule(lr){2-7}

& \multirow{2}{*}{$\sigma=0.10$}
& $f_{\mathrm{std}}$ & \textbf{81.70 $\pm$ 2.39} & \textbf{0.1034 $\pm$ 0.0030} & 100.00 $\pm$ 0.00 & 0.2463 $\pm$ 0.0000 \\
& & $f_{\mathrm{cbm}}$ & 77.20 $\pm$ 2.14 & 0.0918 $\pm$ 0.0038 & 100.00 $\pm$ 0.00 & 0.2463 $\pm$ 0.0000 \\
\cmidrule(lr){2-7}

& \multirow{2}{*}{$\sigma=0.30$}
& $f_{\mathrm{std}}$ & \textbf{62.40 $\pm$ 3.65} & \textbf{0.1263 $\pm$ 0.0058} & 100.00 $\pm$ 0.00 & 0.7376 $\pm$ 0.0004 \\
& & $f_{\mathrm{cbm}}$ & 57.10 $\pm$ 4.57 & 0.1183 $\pm$ 0.0045 & 100.00 $\pm$ 0.00 & 0.7375 $\pm$ 0.0003 \\

\midrule
\midrule

\multirow{6}{*}{Concept RM}
& \multirow{2}{*}{$\rho=0.05$}
& $f_{\mathrm{std}}$ & 100.00 $\pm$ 0.00 & \textbf{0.002 $\pm$ 0.003} & 100.00 $\pm$ 0.00 & 1.000 $\pm$ 0.000 \\
& & $f_{\mathrm{cbm}}$ & 100.00 $\pm$ 0.00 & 0.001 $\pm$ 0.002 & 100.00 $\pm$ 0.00 & 1.000 $\pm$ 0.000 \\
\cmidrule(lr){2-7}

& \multirow{2}{*}{$\rho=0.10$}
& $f_{\mathrm{std}}$ & 95.10 $\pm$ 0.55 & \textbf{0.002 $\pm$ 0.003} & 100.00 $\pm$ 0.00 & 1.834 $\pm$ 0.014 \\
& & $f_{\mathrm{cbm}}$ & \textbf{95.80 $\pm$ 0.97} & 0.001 $\pm$ 0.002 & 100.00 $\pm$ 0.00 & 1.822 $\pm$ 0.013 \\
\cmidrule(lr){2-7}

& \multirow{2}{*}{$\rho=0.20$}
& $f_{\mathrm{std}}$ & \textbf{64.50 $\pm$ 2.81} & \textbf{0.009 $\pm$ 0.003} & 100.00 $\pm$ 0.00 & 3.000 $\pm$ 0.031 \\
& & $f_{\mathrm{cbm}}$ & 63.40 $\pm$ 4.08 & 0.002 $\pm$ 0.004 & 100.00 $\pm$ 0.00 & 3.084 $\pm$ 0.024 \\

\bottomrule
\end{tabular}

\caption{
Unified randomized smoothing results in latent and concept spaces.
Latent RM uses Gaussian noise $\delta\sim\mathcal N(0,\sigma^2 I)$
and reports certified $\ell_2$ radius $R$.
Concept RM uses independent Bernoulli concept flips with probability
$\rho$ and reports certified Hamming radius $\Delta$.
SmAcc measures agreement of non-abstaining smoothed predictions
with the shared clean reference labels, counting abstentions as
incorrect. Certified radii are averaged over correctly classified,
non-abstained samples. Values are mean $\pm$ standard deviation
over five random subsets of 200 images.
Bold indicates the better value between $f_{\mathrm{std}}$ and
$f_{\mathrm{cbm}}$ for CUB-200-2011 only;
we omit highlighting on RIVAL-10.
}
\label{tab:rm_merged}
\end{table*}

\autoref{tab:rm_merged} reports the complete set of randomized smoothing experimental results. Under latent-space smoothing,
the \textit{ResNet baseline generally achieves higher SmAcc and larger certified radii
than the PCBM} on CUB-200-2011, especially as $\sigma$ increases. This indicates
that, on generator native inputs, the concept bottleneck does not improve
certifiable robustness to continuous latent perturbations. On the original
RIVAL-10 split, both models are saturated: SmAcc remains at 100\% and certified
radii are nearly identical across all $\sigma$, reflecting the strong class
separability of this benchmark and
indicating that samples remain far from decision boundaries in latent space.


\subsection{Robustness under Task Conditions}

\subsubsection{Effect of Class Structure and Semantic Similarity}

\begin{figure*}
    \centering
    \begin{tabular}{c}
      \includegraphics[width=0.85\linewidth]{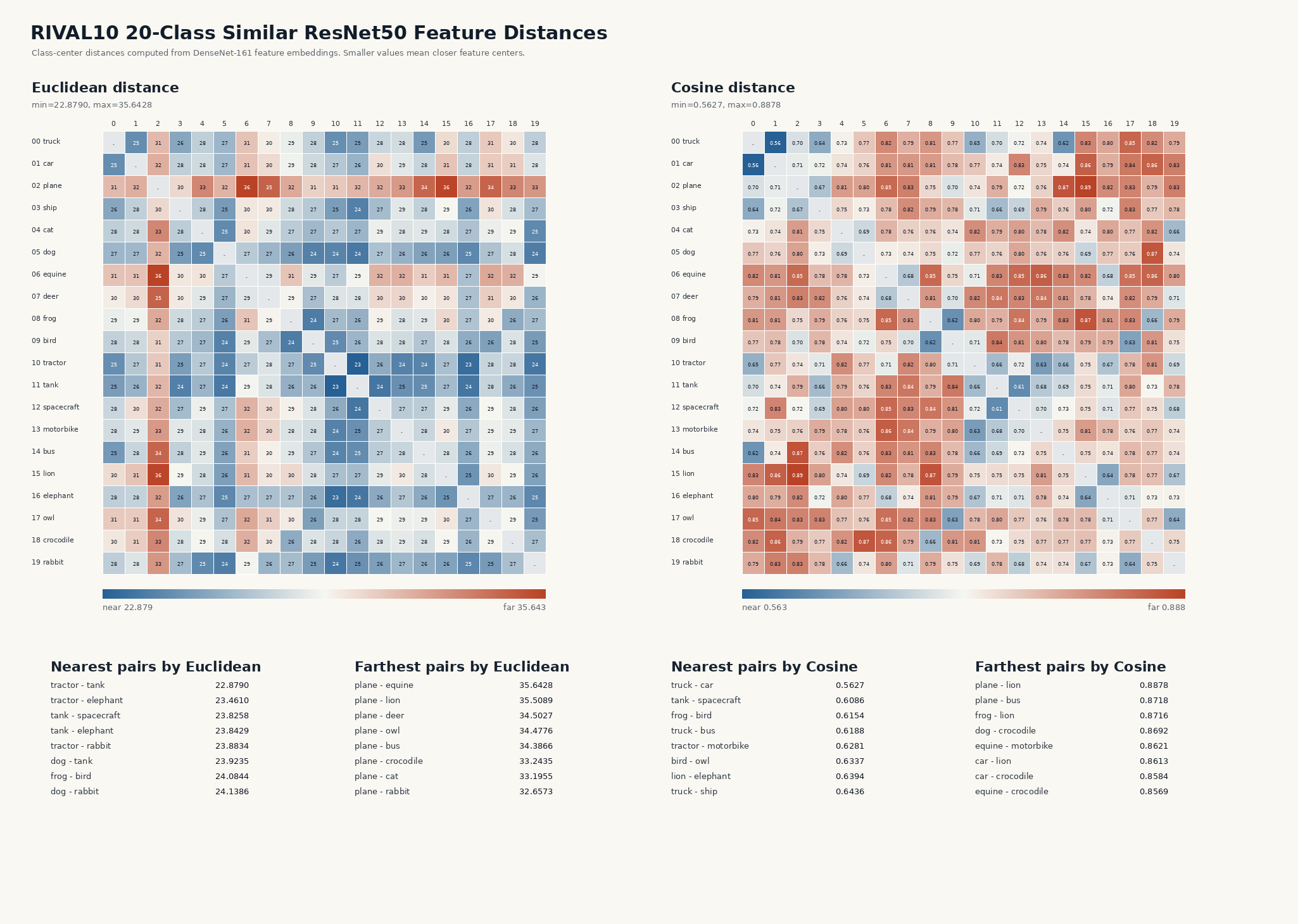}  \\  
       \includegraphics[width=0.85\linewidth]{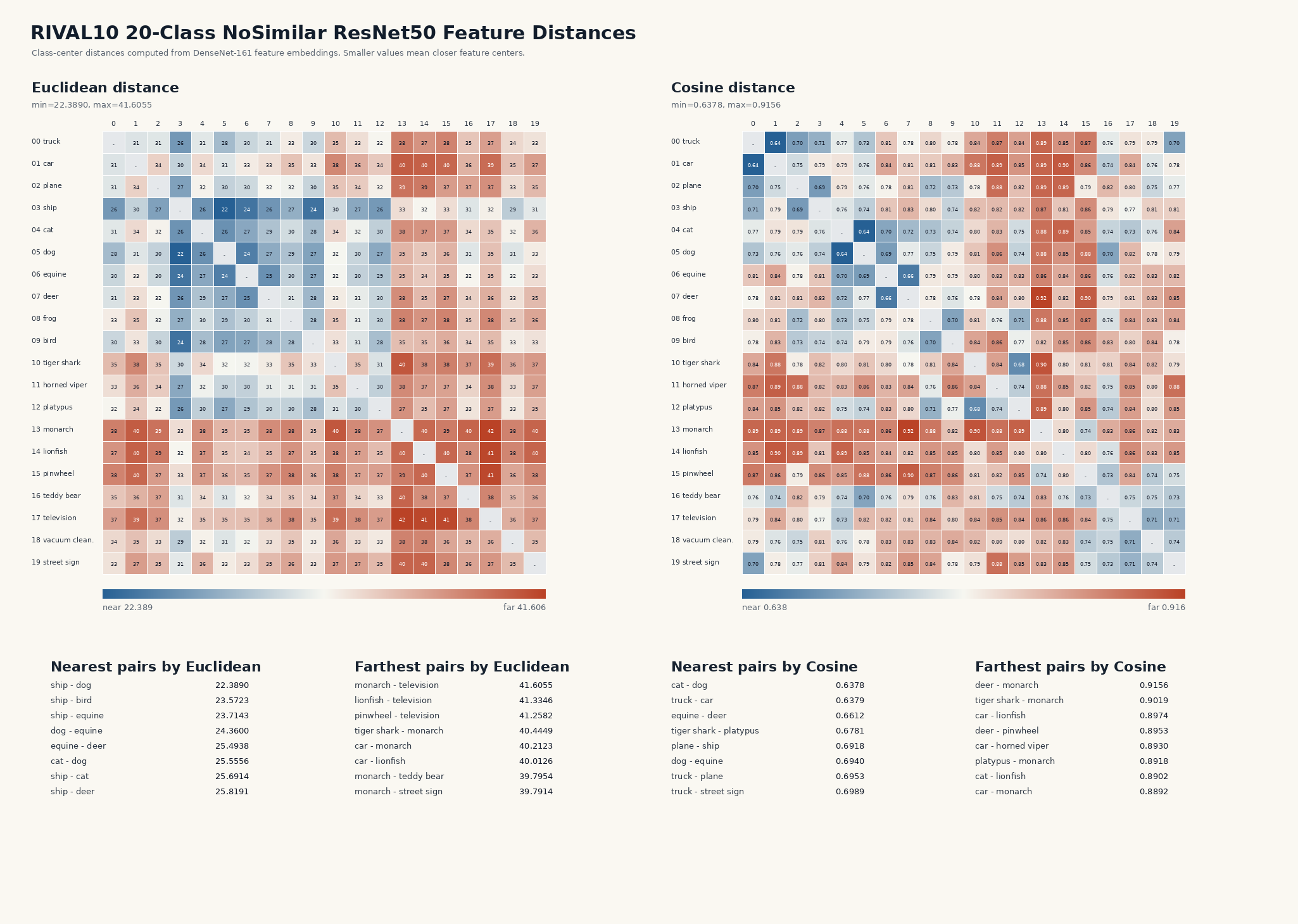} 
    \end{tabular}
    
    \caption{\textbf{Distance heatmap among 20 classes under similar and dissimilar settings.} Class distances are computed as Euclidean distances between class‑level centroid features extracted using a ResNet‑50 backbone.}
    \label{fig:class_dis}
\end{figure*}

To construct additional classes for RIVAL‑10, we use a large language model to identify two sets of ImageNet classes: ten classes that are semantically similar to the original RIVAL‑10 categories and ten classes that are semantically dissimilar. The selected classes are shown in \autoref{fig:class_dis}. Class similarity is estimated using feature distances extracted from a ResNet‑50 backbone: for each class, we compute a centroid feature via K-means clustering and measure pairwise similarity using Euclidean distances between these class centroids.

\autoref{tab:combined_classes_geo_sem} and \autoref{tab:rival_ablation_rm_merged} present the complete set of empirical and certified robustness experiments.

\begin{table*}[h]
\centering
\small
\setlength{\tabcolsep}{1.8pt}
\caption{Robustness on RIVAL-10 (\textbf{18 concepts, varying class count}). Backbone: ResNet50. \textbf{Geo.}: latent $\ell_2$ perturbation (values mean\,$\pm$\,std). \textbf{Sem.}: concept number flip, $\tau$ concepts flipped (global average). CFR\,(count): concept flip rate / $\tau$; PFR\,(\%): prediction flip rate; DFS: mean PCBM concept distance.}
\label{tab:combined_classes_geo_sem}
\resizebox{\linewidth}{!}{%
\begin{tabular}{l r |c c c c | c c c c | c c c c | c c c c}
\toprule
\multicolumn{2}{c|}{\textbf{Perturbation}}
    & \multicolumn{4}{c|}{\textbf{ 10+ 5 dissimilar classes }}
    & \multicolumn{4}{c|}{\textbf{ 10+ 5 similar classes }}
    & \multicolumn{4}{c|}{\textbf{ 10+ 10 dissimilar classes }}
    & \multicolumn{4}{c}{\textbf{ 10+ 10 similar classes }} \\
\cmidrule(lr){1-2}\cmidrule(lr){3-6}\cmidrule(lr){7-10}\cmidrule(lr){11-14}\cmidrule(lr){15-18}
Type & Budget
    & CFR & \multicolumn{2}{c}{PFR\,(\%)\,$\downarrow$} & DFS\,$\downarrow$
    & CFR & \multicolumn{2}{c}{PFR\,(\%)\,$\downarrow$} & DFS\,$\downarrow$
    & CFR & \multicolumn{2}{c}{PFR\,(\%)\,$\downarrow$} & DFS\,$\downarrow$
    & CFR & \multicolumn{2}{c}{PFR\,(\%)\,$\downarrow$} & DFS\,$\downarrow$ \\
\cmidrule(lr){4-5}\cmidrule(lr){8-9}\cmidrule(lr){12-13}\cmidrule(lr){16-17}
 & & & $f_\mathrm{std}$ & $f_\mathrm{cbm}$ & & & $f_\mathrm{std}$ & $f_\mathrm{cbm}$ & & & $f_\mathrm{std}$ & $f_\mathrm{cbm}$ & & & $f_\mathrm{std}$ & $f_\mathrm{cbm}$ & \\
\midrule
  \multirow{5}{*}{Geo.} & $\varepsilon=0.05$ &  0.81 & $0.29\pm0.16$ & $0.06\pm0.04$ & $0.37\pm0.01$ & 0.56 & $3.72\pm0.64$ & $2.06\pm0.47$ & $0.60\pm0.01$ & 0.44 & $0.06\pm0.04$ & $0.08\pm0.05$ & $0.48\pm0.01$ & 0.14 & $1.86\pm0.35$ & $1.85\pm0.43$ & $0.45\pm0.01$ \\
   & $\varepsilon=0.1$  &  1.77 & $0.46\pm0.19$ & $0.15\pm0.08$ & $0.60\pm0.02$ & 1.11 & $5.96\pm0.86$ & $3.61\pm0.65$ & $0.92\pm0.02$ & 0.93 & $0.17\pm0.08$ & $0.25\pm0.11$ & $0.75\pm0.02$ & 0.18 & $3.81\pm0.57$ & $3.40\pm0.65$ & $0.72\pm0.02$ \\
   & $\varepsilon=0.3$  &  5.77 & $0.80\pm0.24$ & $0.33\pm0.11$ & $1.10\pm0.03$ & 3.85 & $10.06\pm1.20$ & $6.94\pm0.98$ & $1.60\pm0.03$ & 3.38 & $0.47\pm0.14$ & $0.61\pm0.17$ & $1.32\pm0.03$ & 1.02 & $8.01\pm0.98$ & $6.10\pm0.93$ & $1.29\pm0.03$ \\
   & $\varepsilon=0.5$  &  9.83 & $1.03\pm0.29$ & $0.53\pm0.16$ & $1.39\pm0.04$ & 7.17 & $11.71\pm1.33$ & $8.34\pm1.11$ & $1.95\pm0.04$ & 6.89 & $0.59\pm0.16$ & $0.81\pm0.18$ & $1.67\pm0.03$ & 2.42 & $9.91\pm1.12$ & $7.37\pm1.03$ & $1.62\pm0.03$ \\
   & $\varepsilon=0.8$  &  15.73 & $1.33\pm0.36$ & $0.83\pm0.23$ & $1.69\pm0.05$ & 12.37 & $12.80\pm1.41$ & $9.55\pm1.22$ & $2.31\pm0.04$ & 13.04 & $0.78\pm0.16$ & $1.23\pm0.21$ & $2.05\pm0.04$ & 5.86 & $11.59\pm1.19$ & $8.63\pm1.10$ & $1.99\pm0.04$ \\
\midrule
  \multirow{5}{*}{Sem.}
& $\tau=1$
& 1 & 0.99 $\pm$ 0.22 & 0.46 $\pm$ 0.14 & 1.40 $\pm$ 0.01
& 1 & 11.07 $\pm$ 0.89 & 7.87 $\pm$ 1.09 & 1.92 $\pm$ 0.03
& 1 & 0.29 $\pm$ 0.11 & 0.56 $\pm$ 0.11 & 1.69 $\pm$ 0.03
& 1 & 9.64 $\pm$ 1.72 & 6.93 $\pm$ 1.67 & 1.62 $\pm$ 0.05 \\

& $\tau=2$
& 2 & 1.28 $\pm$ 0.29 & 0.77 $\pm$ 0.18 & 1.89 $\pm$ 0.05
& 2 & 12.70 $\pm$ 1.31 & 9.34 $\pm$ 1.10 & 2.33 $\pm$ 0.04
& 2 & 0.78 $\pm$ 0.20 & 1.36 $\pm$ 0.18 & 2.19 $\pm$ 0.03
& 2 & 9.72 $\pm$ 0.69 & 7.03 $\pm$ 0.31 & 1.89 $\pm$ 0.02 \\

& $\tau=5$
& 5 & 1.09 $\pm$ 0.22 & 0.89 $\pm$ 0.15 & 2.95 $\pm$ 0.03
& 5 & 13.76 $\pm$ 1.17 & 10.48 $\pm$ 1.17 & 3.11 $\pm$ 0.04
& 5 & 6.77 $\pm$ 0.31 & 7.66 $\pm$ 0.39 & 3.48 $\pm$ 0.05
& 5 & 13.73 $\pm$ 1.13 & 9.61 $\pm$ 1.29 & 2.50 $\pm$ 0.04 \\

& $\tau=7$
& 7 & 1.93 $\pm$ 0.15 & 2.34 $\pm$ 0.12 & 3.76 $\pm$ 0.07
& 7 & 16.01 $\pm$ 0.73 & 13.05 $\pm$ 0.63 & 3.67 $\pm$ 0.03
& 7 & 15.08 $\pm$ 0.32 & 17.03 $\pm$ 0.49 & 4.56 $\pm$ 0.07
& 7 & 17.65 $\pm$ 1.40 & 13.14 $\pm$ 1.31 & 3.02 $\pm$ 0.07 \\

& $\tau=10$
& 10 & 6.32 $\pm$ 0.63 & 8.27 $\pm$ 0.82 & 5.42 $\pm$ 0.16
& 10 & 20.91 $\pm$ 1.04 & 18.07 $\pm$ 0.99 & 4.78 $\pm$ 0.05
& 10 & 34.66 $\pm$ 1.34 & 37.86 $\pm$ 1.26 & 6.59 $\pm$ 0.07
& 10 & 21.97 $\pm$ 2.56 & 17.85 $\pm$ 2.26 & 3.76 $\pm$ 0.12 \\
\bottomrule
\end{tabular}
}
\end{table*}
\begin{table*}[t]
\centering
\small
\setlength{\tabcolsep}{2.6pt}
\resizebox{\textwidth}{!}{%
\begin{tabular}{l c c | c c | c c | c c | c c}
\toprule
\textbf{Type} & \textbf{Budget} & \textbf{Model}
& \multicolumn{2}{c|}{\textbf{10 + 5 dissimilar classes}}
& \multicolumn{2}{c|}{\textbf{10 + 5 similar classes}}
& \multicolumn{2}{c|}{\textbf{10 + 10 dissimilar classes}}
& \multicolumn{2}{c}{\textbf{10 + 10 similar classes}} \\
\cmidrule(lr){4-5}\cmidrule(lr){6-7}\cmidrule(lr){8-9}\cmidrule(lr){10-11}
& & & SmAcc (\%) $\uparrow$ & $R/\Delta^\star\uparrow$
& SmAcc (\%) $\uparrow$ & $R/\Delta^\star\uparrow$
& SmAcc (\%) $\uparrow$ & $R/\Delta^\star\uparrow$
& SmAcc (\%) $\uparrow$ & $R/\Delta^\star\uparrow$ \\
\midrule
\multirow{8}{*}{Latent RM}
& $\sigma=0.01$ & $f_{\mathrm{std}}$ & 100.00 $\pm$ 0.00 & 0.0248 $\pm$ 0.0001 & 98.50 $\pm$ 0.35 & 0.0240 $\pm$ 0.0002 & 100.00 $\pm$ 0.00 & 0.0247 $\pm$ 0.0000 & 99.90 $\pm$ 0.22 & 0.0240 $\pm$ 0.0002 \\
& & $f_{\mathrm{cbm}}$ & 100.00 $\pm$ 0.00 & 0.0249 $\pm$ 0.0000 & 99.90 $\pm$ 0.22 & 0.0242 $\pm$ 0.0001 & 100.00 $\pm$ 0.00 & 0.0247 $\pm$ 0.0000 & 99.80 $\pm$ 0.27 & 0.0242 $\pm$ 0.0001 \\
\cmidrule(lr){2-11}
& $\sigma=0.05$ & $f_{\mathrm{std}}$ & 99.80 $\pm$ 0.27 & 0.1225 $\pm$ 0.0004 & 95.10 $\pm$ 0.82 & 0.1145 $\pm$ 0.0012 & 100.00 $\pm$ 0.00 & 0.1224 $\pm$ 0.0002 & 97.80 $\pm$ 0.76 & 0.1119 $\pm$ 0.0008 \\
& & $f_{\mathrm{cbm}}$ & 100.00 $\pm$ 0.00 & 0.1236 $\pm$ 0.0003 & 96.80 $\pm$ 0.45 & 0.1178 $\pm$ 0.0004 & 100.00 $\pm$ 0.00 & 0.1220 $\pm$ 0.0004 & 96.70 $\pm$ 0.27 & 0.1176 $\pm$ 0.0008 \\
\cmidrule(lr){2-11}
& $\sigma=0.10$ & $f_{\mathrm{std}}$ & 99.80 $\pm$ 0.27 & 0.2430 $\pm$ 0.0012 & 90.60 $\pm$ 0.89 & 0.2318 $\pm$ 0.0031 & 100.00 $\pm$ 0.00 & 0.2425 $\pm$ 0.0004 & 95.30 $\pm$ 1.20 & 0.2192 $\pm$ 0.0041 \\
& & $f_{\mathrm{cbm}}$ & 100.00 $\pm$ 0.00 & 0.2459 $\pm$ 0.0009 & 95.10 $\pm$ 0.82 & 0.2325 $\pm$ 0.0017 & 100.00 $\pm$ 0.00 & 0.2404 $\pm$ 0.0008 & 95.30 $\pm$ 0.57 & 0.2332 $\pm$ 0.0011 \\
\cmidrule(lr){2-11}
& $\sigma=0.30$ & $f_{\mathrm{std}}$ & 99.40 $\pm$ 0.42 & 0.7219 $\pm$ 0.0041 & 85.80 $\pm$ 0.84 & 0.7015 $\pm$ 0.0061 & 100.00 $\pm$ 0.00 & 0.7092 $\pm$ 0.0015 & 88.00 $\pm$ 0.35 & 0.6584 $\pm$ 0.0076 \\
& & $f_{\mathrm{cbm}}$ & 100.00 $\pm$ 0.00 & 0.7277 $\pm$ 0.0051 & 90.80 $\pm$ 0.57 & 0.6958 $\pm$ 0.0070 & 100.00 $\pm$ 0.00 & 0.6925 $\pm$ 0.0052 & 91.50 $\pm$ 0.61 & 0.6820 $\pm$ 0.0057 \\
\midrule
\multirow{6}{*}{Concept RM}
& $\rho=0.05$ & $f_{\mathrm{std}}$ & 100.00 $\pm$ 0.00 & 0.937 $\pm$ 0.014 & 94.40 $\pm$ 0.65 & 0.862 $\pm$ 0.016 & 100.00 $\pm$ 0.00 & 0.941 $\pm$ 0.016 & 98.30 $\pm$ 0.45 & 0.761 $\pm$ 0.026 \\
& & $f_{\mathrm{cbm}}$ & 100.00 $\pm$ 0.00 & 0.960 $\pm$ 0.009 & 97.80 $\pm$ 0.67 & 0.859 $\pm$ 0.006 & 100.00 $\pm$ 0.00 & 0.889 $\pm$ 0.012 & 97.70 $\pm$ 0.45 & 0.866 $\pm$ 0.013 \\
\cmidrule(lr){2-11}
& $\rho=0.10$ & $f_{\mathrm{std}}$ & 100.00 $\pm$ 0.00 & 0.954 $\pm$ 0.010 & 88.30 $\pm$ 1.44 & 0.930 $\pm$ 0.017 & 99.80 $\pm$ 0.27 & 0.957 $\pm$ 0.019 & 89.60 $\pm$ 0.65 & 0.829 $\pm$ 0.024 \\
& & $f_{\mathrm{cbm}}$ & 100.00 $\pm$ 0.00 & 0.967 $\pm$ 0.010 & 91.30 $\pm$ 1.30 & 0.934 $\pm$ 0.008 & 100.00 $\pm$ 0.00 & 0.904 $\pm$ 0.013 & 92.70 $\pm$ 0.76 & 0.913 $\pm$ 0.010 \\
\cmidrule(lr){2-11}
& $\rho=0.20$ & $f_{\mathrm{std}}$ & 100.00 $\pm$ 0.00 & 2.832 $\pm$ 0.053 & 86.00 $\pm$ 1.77 & 2.478 $\pm$ 0.045 & 99.40 $\pm$ 0.22 & 1.529 $\pm$ 0.031 & 86.60 $\pm$ 0.96 & 2.600 $\pm$ 0.118 \\
& & $f_{\mathrm{cbm}}$ & 100.00 $\pm$ 0.00 & 2.799 $\pm$ 0.040 & 88.90 $\pm$ 2.13 & 2.511 $\pm$ 0.045 & 99.80 $\pm$ 0.27 & 1.339 $\pm$ 0.028 & 90.60 $\pm$ 0.55 & 2.575 $\pm$ 0.077 \\
\bottomrule
\end{tabular}%
}
\caption{Randomized smoothing ablation on RIVAL-10 variants with 18 concepts. Latent RM reports the certified Gaussian latent $\ell_2$ radius $R$; Concept RM reports the certified Bernoulli Hamming radius $\Delta^\star$. SmAcc is the fraction of non-abstaining smoothed predictions matching the shared clean reference label, with abstentions counted as incorrect. Radii are averaged over correctly classified, non-abstained inputs. Values are mean $\pm$ standard deviation over five random subsets of 200 images. Gaussian and Bernoulli results use $N=1000$ and simultaneous confidence bounds at $1-\alpha=95\%$.}
\label{tab:rival_ablation_rm_merged}
\end{table*}

\subsubsection{Effect of Concept Vocabulary Size}

\autoref{tab:empirical_concept_count_merged} and
\autoref{tab:rival10_10c_27_36_latent_concept_corrected} presents the complete set of empirical and certified robustness experiments.

\begin{table*}[h]
\centering
\small
\setlength{\tabcolsep}{2.2pt}
\caption{
Empirical robustness under geometric (latent $\ell_2$) and semantic (concept flip)
perturbations on RIVAL-10 with varying concept vocabulary size.
CFR: concept flip rate; PFR (\%): prediction flip rate; DFS: PCBM concept distance.
}
\label{tab:empirical_concept_count_merged}
\resizebox{\linewidth}{!}{%
\begin{tabular}{l r | c c c c | c c c c | c c c c}
\toprule
\multicolumn{2}{c|}{\textbf{Perturbation}}
& \multicolumn{4}{c|}{\textbf{18 concepts}}
& \multicolumn{4}{c|}{\textbf{27 concepts}}
& \multicolumn{4}{c}{\textbf{36 concepts}} \\
\cmidrule(lr){1-2}
\cmidrule(lr){3-6}
\cmidrule(lr){7-10}
\cmidrule(lr){11-14}
Type & Budget
& CFR & \multicolumn{2}{c}{PFR (\%) $\downarrow$} & DFS $\downarrow$
& CFR & \multicolumn{2}{c}{PFR (\%) $\downarrow$} & DFS $\downarrow$
& CFR & \multicolumn{2}{c}{PFR (\%) $\downarrow$} & DFS $\downarrow$ \\
\cmidrule(lr){4-5}\cmidrule(lr){8-9}\cmidrule(lr){12-13}
& &
& $f_{\mathrm{std}}$ & $f_{\mathrm{cbm}}$ &
& & $f_{\mathrm{std}}$ & $f_{\mathrm{cbm}}$ &
& & $f_{\mathrm{std}}$ & $f_{\mathrm{cbm}}$ & \\
\midrule

\multirow{5}{*}{Geo.}
& $\varepsilon=0.05$
& 0.32 & 0.00 & 0.00 & $0.48\pm0.01$
& 0.15 & 0.00 & 0.00 & $0.71\pm0.02$
& 1.0 & 0.00 & 0.00 & $0.77\pm0.03$ \\

& $\varepsilon=0.10$
& 0.67 & 0.00 & 0.00 & $0.74\pm0.01$
& 0.56 & 0.00 & $0.01\pm0.01$ & $1.10\pm0.02$
& 1.67 & 0.00 & 0.00 & $1.21\pm0.04$ \\

& $\varepsilon=0.30$
& 2.11 & 0.00 & 0.00 & $1.29\pm0.02$
& 2.58 & 0.00 & $0.01\pm0.01$ & $1.91\pm0.04$
& 4.09 & 0.00 & 0.00 & $2.20\pm0.06$ \\

& $\varepsilon=0.50$
& 4.09 & 0.00 & 0.00 & $1.61\pm0.03$
& 5.16 & $0.01\pm0.00$ & $0.01\pm0.01$ & $2.36\pm0.04$
& 6.94 & $0.01\pm0.00$ & 0.00 & $2.86\pm0.08$ \\

& $\varepsilon=0.80$
& 8.36 & $0.02\pm0.01$ & $0.03\pm0.02$ & $1.95\pm0.03$
& 9.92 & $0.04\pm0.01$ & $0.04\pm0.01$ & $2.85\pm0.05$
& 11.97 & $0.04\pm0.01$ & $0.06\pm0.01$ & $3.67\pm0.10$ \\

\midrule

\multirow{4}{*}{Sem.}
& $\tau=1$
& 1 & 0.00 $\pm$ 0.00 & 0.00 $\pm$ 0.00 & 1.82 $\pm$ 0.02
& 1 & 0.03 $\pm$ 0.02 & 0.04 $\pm$ 0.02 & 2.38 $\pm$ 0.06
& 1 & 0.01 $\pm$ 0.01 & 0.00 $\pm$ 0.00 & 2.25 $\pm$ 0.05 \\

& $\tau=2$
& 2 & 0.07 $\pm$ 0.02 & 0.09 $\pm$ 0.02 & 2.29 $\pm$ 0.05
& 2 & 0.16 $\pm$ 0.05 & 0.18 $\pm$ 0.04 & 3.03 $\pm$ 0.06
& 2 & 0.02 $\pm$ 0.01 & 0.02 $\pm$ 0.01 & 3.05 $\pm$ 0.07 \\

& $\tau=5$
& 5 & 0.75 $\pm$ 0.07 & 0.77 $\pm$ 0.07 & 3.17 $\pm$ 0.05
& 5 & 1.42 $\pm$ 0.20 & 1.44 $\pm$ 0.20 & 4.73 $\pm$ 0.07
& 5 & 0.29 $\pm$ 0.05 & 0.36 $\pm$ 0.05 & 4.67 $\pm$ 0.17 \\

& $\tau=10$
& 10 & 5.79 $\pm$ 0.52 & 5.74 $\pm$ 0.52 & 5.13 $\pm$ 0.10
& 10 & 9.42 $\pm$ 0.99 & 9.29 $\pm$ 0.88 & 9.10 $\pm$ 0.24
& 10 & 2.65 $\pm$ 0.18 & 3.11 $\pm$ 0.19 & 7.99 $\pm$ 0.14 \\

\bottomrule
\end{tabular}%
}
\end{table*}

\begin{table*}[t]
\centering
\small
\setlength{\tabcolsep}{2pt}
\caption{RIVAL-10 10-class randomized smoothing with 27 and 36 concepts (ResNet50). Latent RM reports Gaussian certified $\ell_2$ radius $R$, and Concept RM reports Bernoulli certified Hamming radius $\Delta^\star$. SmAcc measures agreement with shared clean reference labels, with abstentions counted as incorrect; radii are averaged over correct, non-abstaining inputs. Values are mean $\pm$ standard deviation over five 200-image subsets ($N=1000$, $\alpha=0.05$).}
\label{tab:rival10_10c_27_36_latent_concept_corrected}
\begin{tabular}{lllcccc}
\toprule
Type & Budget & Model & \multicolumn{2}{c}{27 concepts} & \multicolumn{2}{c}{36 concepts} \\
\cmidrule(lr){4-5}\cmidrule(lr){6-7}
& & & SmAcc (\%) $\uparrow$ & $R/\Delta^\star\uparrow$ & SmAcc (\%) $\uparrow$ & $R/\Delta^\star\uparrow$ \\
\midrule
\multirow{8}{*}{Latent RM}
& $\sigma=0.01$ & $f_{\mathrm{std}}$ & 100.00 $\pm$ 0.00 & 0.0251 $\pm$ 0.0000 & 100.00 $\pm$ 0.00 & 0.0251 $\pm$ 0.0000 \\
& & $f_{\mathrm{cbm}}$ & 100.00 $\pm$ 0.00 & 0.0251 $\pm$ 0.0000 & 100.00 $\pm$ 0.00 & 0.0251 $\pm$ 0.0000 \\
\cmidrule(lr){2-7}
& $\sigma=0.05$ & $f_{\mathrm{std}}$ & 100.00 $\pm$ 0.00 & 0.1257 $\pm$ 0.0000 & 100.00 $\pm$ 0.00 & 0.1256 $\pm$ 0.0001 \\
& & $f_{\mathrm{cbm}}$ & 100.00 $\pm$ 0.00 & 0.1256 $\pm$ 0.0001 & 100.00 $\pm$ 0.00 & 0.1256 $\pm$ 0.0001 \\
\cmidrule(lr){2-7}
& $\sigma=0.10$ & $f_{\mathrm{std}}$ & 100.00 $\pm$ 0.00 & 0.2513 $\pm$ 0.0000 & 100.00 $\pm$ 0.00 & 0.2511 $\pm$ 0.0002 \\
& & $f_{\mathrm{cbm}}$ & 100.00 $\pm$ 0.00 & 0.2511 $\pm$ 0.0002 & 100.00 $\pm$ 0.00 & 0.2512 $\pm$ 0.0002 \\
\cmidrule(lr){2-7}
& $\sigma=0.30$ & $f_{\mathrm{std}}$ & 100.00 $\pm$ 0.00 & 0.7511 $\pm$ 0.0004 & 100.00 $\pm$ 0.00 & 0.7518 $\pm$ 0.0010 \\
& & $f_{\mathrm{cbm}}$ & 100.00 $\pm$ 0.00 & 0.7498 $\pm$ 0.0004 & 100.00 $\pm$ 0.00 & 0.7513 $\pm$ 0.0009 \\
\midrule
\multirow{6}{*}{Concept RM}
& $\rho=0.05$ & $f_{\mathrm{std}}$ & 100.00 $\pm$ 0.00 & 0.983 $\pm$ 0.003 & 100.00 $\pm$ 0.00 & 0.996 $\pm$ 0.002 \\
& & $f_{\mathrm{cbm}}$ & 100.00 $\pm$ 0.00 & 0.981 $\pm$ 0.004 & 100.00 $\pm$ 0.00 & 0.996 $\pm$ 0.002 \\
\cmidrule(lr){2-7}
& $\rho=0.10$ & $f_{\mathrm{std}}$ & 100.00 $\pm$ 0.00 & 1.449 $\pm$ 0.015 & 100.00 $\pm$ 0.00 & 1.657 $\pm$ 0.017 \\
& & $f_{\mathrm{cbm}}$ & 100.00 $\pm$ 0.00 & 1.451 $\pm$ 0.005 & 100.00 $\pm$ 0.00 & 1.615 $\pm$ 0.032 \\
\cmidrule(lr){2-7}
& $\rho=0.20$ & $f_{\mathrm{std}}$ & 100.00 $\pm$ 0.00 & 2.002 $\pm$ 0.038 & 100.00 $\pm$ 0.00 & 2.152 $\pm$ 0.016 \\
& & $f_{\mathrm{cbm}}$ & 100.00 $\pm$ 0.00 & 2.011 $\pm$ 0.047 & 100.00 $\pm$ 0.00 & 2.200 $\pm$ 0.019 \\
\bottomrule
\end{tabular}
\end{table*}

\subsubsection{Experiments on DenseNet-161.}
We test similar experiments on DenseNet-161~\citep{huang2017densely} and the results are shown in \autoref{tab:densenet_direct} and \autoref{tab:densenet_gaussian}.

\begin{table*}[thbp]
  \centering
  \small
  \setlength{\tabcolsep}{1.8pt}
  \resizebox{\textwidth}{!}{%
  \begin{tabular}{l l | c c c | c c c | c c c | c c c}
  \toprule
  & & \multicolumn{3}{c|}{\textbf{10 class;K=18}}
    & \multicolumn{3}{c|}{\textbf{10 + 10 similar classes; K=18}}
    & \multicolumn{3}{c|}{\textbf{10 + 10 dissimilar classes; K=18}}
    & \multicolumn{3}{c}{\textbf{10 + 10 dissimilar classes; K=27}} \\
  \cmidrule(lr){3-5}\cmidrule(lr){6-8}\cmidrule(lr){9-11}\cmidrule(lr){12-14}
  \textbf{Type} & \textbf{Budget}
    & PFR$_{f_{\text{dn}}}$ & PFR$_{f_{\text{cbm}}}$ & DFS
    & PFR$_{f_{\text{dn}}}$ & PFR$_{f_{\text{cbm}}}$ & DFS
    & PFR$_{f_{\text{dn}}}$ & PFR$_{f_{\text{cbm}}}$ & DFS
    & PFR$_{f_{\text{dn}}}$ & PFR$_{f_{\text{cbm}}}$ & DFS \\
  \midrule
  \multicolumn{2}{l}{\textit{Geo.\ ($\ell_2$ latent)}} & & & & & & & & & & & & \\
    & $\varepsilon=0.05$ & $0.00 \pm 0.00$ & $0.00$ & $0.391 \pm 0.007$ & $1.72 \pm
  0.41$ & $4.21 \pm 0.61$ & $0.537 \pm 0.004$ & $0.03 \pm 0.04$ & $0.07 \pm 0.05$ &
  $0.491 \pm 0.012$ & $0.22 \pm 0.09$ & $0.30 \pm 0.18$ & $0.710 \pm 0.013$ \\
    & $\varepsilon=0.1$ & $0.00$ & $0.00 \pm 0.00$ & $0.663 \pm 0.012$ & $3.39 \pm 0.52$
   & $6.02 \pm 0.75$ & $0.895 \pm 0.023$ & $0.10 \pm 0.05$ & $0.25 \pm 0.13$ & $0.830
  \pm 0.024$ & $0.49 \pm 0.17$ & $0.36 \pm 0.22$ & $1.192 \pm 0.035$ \\
    & $\varepsilon=0.3$ & $0.09 \pm 0.03$ & $0.00 \pm 0.00$ & $1.343 \pm 0.010$ & $6.68
  \pm 1.04$ & $9.95 \pm 1.81$ & $1.775 \pm 0.044$ & $0.12 \pm 0.07$ & $0.29 \pm 0.15$ &
  $1.547 \pm 0.037$ & $0.66 \pm 0.17$ & $0.30 \pm 0.14$ & $2.230 \pm 0.061$ \\
    & $\varepsilon=0.5$ & $0.10 \pm 0.06$ & $0.01 \pm 0.00$ & $1.761 \pm 0.021$ & $9.07
  \pm 1.25$ & $12.24 \pm 1.70$ & $2.341 \pm 0.064$ & $0.18 \pm 0.08$ & $0.43 \pm 0.15$ &
   $2.054 \pm 0.049$ & $1.04 \pm 0.15$ & $0.43 \pm 0.08$ & $2.879 \pm 0.085$ \\
    & $\varepsilon=0.8$ & $0.16 \pm 0.05$ & $0.01 \pm 0.01$ & $2.219 \pm 0.025$ & $8.51
  \pm 0.80$ & $11.51 \pm 1.04$ & $2.913 \pm 0.059$ & $0.29 \pm 0.07$ & $0.51 \pm 0.07$ &
   $2.588 \pm 0.035$ & $1.24 \pm 0.21$ & $0.45 \pm 0.14$ & $3.594 \pm 0.086$ \\
  \midrule
  \multicolumn{2}{l}{\textit{Sem.\ (Concept flip)}} & & & & & & & & & & & & \\
    & $\tau=1$ & $0.02 \pm 0.01$ & $0.02 \pm 0.02$ & $2.020 \pm 0.019$ & $5.50 \pm 0.87$
   & $9.12 \pm 1.42$ & $2.253 \pm 0.023$ & $0.15 \pm 0.04$ & $0.28 \pm 0.08$ & $2.141
  \pm 0.021$ & $0.94 \pm 0.11$ & $0.30 \pm 0.09$ & $3.054 \pm 0.043$ \\
    & $\tau=2$ & $0.03 \pm 0.01$ & $0.03 \pm 0.01$ & $2.583 \pm 0.017$ & $7.36 \pm 1.32$
   & $11.07 \pm 1.27$ & $2.765 \pm 0.030$ & $0.46 \pm 0.06$ & $0.86 \pm 0.09$ & $2.835
  \pm 0.025$ & $1.59 \pm 0.34$ & $0.43 \pm 0.13$ & $3.965 \pm 0.057$ \\
    & $\tau=5$ & $0.60 \pm 0.05$ & $0.66 \pm 0.04$ & $3.622 \pm 0.041$ & $9.53 \pm 0.93$
   & $14.04 \pm 1.54$ & $3.724 \pm 0.059$ & $5.31 \pm 0.33$ & $7.08 \pm 0.34$ & $4.890
  \pm 0.077$ & $4.13 \pm 0.27$ & $1.96 \pm 0.11$ & $5.781 \pm 0.090$ \\
    & $\tau=7$ & $2.20 \pm 0.21$ & $2.17 \pm 0.24$ & $4.349 \pm 0.023$ & $10.08 \pm
  0.27$ & $16.30 \pm 1.31$ & $4.327 \pm 0.047$ & $14.50 \pm 0.30$ & $17.31 \pm 0.30$ &
  $6.818 \pm 0.060$ & $7.04 \pm 0.46$ & $4.28 \pm 0.16$ & $7.144 \pm 0.051$ \\
    & $\tau=10$ & $7.06 \pm 0.58$ & $6.83 \pm 0.78$ & $5.744 \pm 0.156$ & $14.14 \pm
  0.68$ & $22.17 \pm 0.92$ & $5.264 \pm 0.076$ & $35.51 \pm 1.09$ & $38.38 \pm 1.33$ &
  $10.098 \pm 0.211$ & $14.24 \pm 0.34$ & $10.30 \pm 0.28$ & $9.318 \pm 0.087$ \\
  \bottomrule
  \end{tabular}}
  \caption{DenseNet-161 direct robustness on RIVAL-10 variants. PFR\,(\%): prediction
  flip rate for $f_{\text{dn}}$ (DenseNet-161) and $f_{\text{cbm}}$; DFS: average PCBM
  concept distance. Mean\,$\pm$\,std over 5 random subsets of 200 images.}
  \label{tab:densenet_direct}
  \end{table*}

\begin{table*}[thbp]
\centering
\small
\setlength{\tabcolsep}{3pt}
\resizebox{\textwidth}{!}{%
\begin{tabular}{c l | c c | c c | c c}
\toprule
& &
\multicolumn{2}{c|}{\textbf{10 classes; $K=18$}}
& \multicolumn{2}{c|}{\textbf{10+10 similar classes; $K=18$}}
& \multicolumn{2}{c}{\textbf{10+10 dissimilar classes; $K=18$}} \\
\cmidrule(lr){3-4}
\cmidrule(lr){5-6}
\cmidrule(lr){7-8}

$\sigma$ & Model
& SmAcc (\%) $\uparrow$ & $R \uparrow$
& SmAcc (\%) $\uparrow$ & $R \uparrow$
& SmAcc (\%) $\uparrow$ & $R \uparrow$ \\
\midrule

\multirow{2}{*}{0.01}
& $f_{\mathrm{dn}}$
& $100.00\pm0.00$ & $0.0251\pm0.0000$
& $99.70\pm0.27$ & $0.0243\pm0.0002$
& $100.00\pm0.00$ & $0.0247\pm0.0000$ \\

& $f_{\mathrm{cbm}}$
& $100.00\pm0.00$ & $0.0251\pm0.0000$
& $98.60\pm0.74$ & $0.0238\pm0.0002$
& $100.00\pm0.00$ & $0.0247\pm0.0000$ \\
\midrule

\multirow{2}{*}{0.05}
& $f_{\mathrm{dn}}$
& $100.00\pm0.00$ & $0.1255\pm0.0002$
& $97.70\pm1.04$ & $0.1174\pm0.0013$
& $100.00\pm0.00$ & $0.1232\pm0.0003$ \\

& $f_{\mathrm{cbm}}$
& $100.00\pm0.00$ & $0.1257\pm0.0000$
& $93.80\pm1.79$ & $0.1159\pm0.0009$
& $100.00\pm0.00$ & $0.1224\pm0.0005$ \\
\midrule

\multirow{2}{*}{0.10}
& $f_{\mathrm{dn}}$
& $100.00\pm0.00$ & $0.2500\pm0.0008$
& $96.90\pm1.29$ & $0.2294\pm0.0030$
& $100.00\pm0.00$ & $0.2456\pm0.0006$ \\

& $f_{\mathrm{cbm}}$
& $100.00\pm0.00$ & $0.2511\pm0.0001$
& $91.10\pm1.34$ & $0.2313\pm0.0013$
& $100.00\pm0.00$ & $0.2430\pm0.0008$ \\
\midrule

\multirow{2}{*}{0.30}
& $f_{\mathrm{dn}}$
& $100.00\pm0.00$ & $0.7420\pm0.0021$
& $90.40\pm1.08$ & $0.7035\pm0.0045$
& $100.00\pm0.00$ & $0.7261\pm0.0036$ \\

& $f_{\mathrm{cbm}}$
& $100.00\pm0.00$ & $0.7523\pm0.0003$
& $88.50\pm0.79$ & $0.6670\pm0.0061$
& $100.00\pm0.00$ & $0.7091\pm0.0032$ \\

\bottomrule
\end{tabular}%
}

\caption{
DenseNet-161 latent Gaussian randomized smoothing
on RIVAL-10 variants with 18 concepts.
We report smoothed accuracy (SmAcc) and average
certified latent $\ell_2$ radius $R$ over correctly
classified, non-abstained samples.
Values are mean $\pm$ standard deviation over five
random subsets of 200 images.
Certificates use $N=1000$ Monte Carlo samples and
95\% simultaneous Clopper--Pearson confidence bounds.
}
\label{tab:densenet_gaussian}
\end{table*}

  We further evaluate the robustness of DenseNet-161 and its concept-based counterpart on RIVAL-10 variants, with results reported in Tables~\ref{tab:densenet_direct} and~\ref{tab:densenet_gaussian}. Under geometric and semantic perturbations, the relative prediction stability of the two models varies across class configurations. Gaussian randomized smoothing exhibits a similar task-dependent pattern. On the original 10-class setting, both models achieve 100\% smoothed accuracy with nearly identical certified radii. In the 20-class similar setting, DenseNet generally achieves higher smoothed accuracy, while the certified-radius ordering varies with the noise level. In the 20-class dissimilar setting, both models maintain 100\% smoothed accuracy, with DenseNet obtaining slightly larger certified radii at higher noise levels. These results further show that the robustness behavior associated with concept bottlenecks depends on task structure and perturbation conditions, rather than being specific to the ResNet-50 backbone.

\subsubsection{Experiments on Attacks.}

We evaluate adversarial robustness by attacking the entire pipeline using APGD‑CE~\citep{croce2020reliable}; the corresponding results are reported in \autoref{tab:attack_merge_eps}.

\begin{table*}[h]
\centering
\small
\setlength{\tabcolsep}{5pt}
\caption{APGD-CE attack results on the CUB generated dataset based on backbone ResNet-50.}
\label{tab:attack_merge_eps}
\begin{tabular}{l l l r r}
\toprule
Attack-$\epsilon$ & Attack Space & Model & Clean Acc. & Robust Acc. \\
\midrule

\multirow{8}{*}{$0.03$}
& \multirow{4}{*}{latent $z$}
& $f_{std}$& 100\% & 8\% \\
& & \cellcolor{smoothbg}$\hat{f}_{std}$ (smoothed) & \cellcolor{smoothbg}{100}\% & \cellcolor{smoothbg}\textbf{16.5}\% \\
& & $f_{cbm}$ & 100\% & 23.5\% \\
& & \cellcolor{smoothbg}$\hat{f}_{cbm}$ (smoothed) & \cellcolor{smoothbg}100\% & \cellcolor{smoothbg}\textbf{33.5}\% \\
& \multirow{4}{*}{concept $c$}
& $f_{std}$& \textbf{100}\% & 44\% \\
& & \cellcolor{smoothbg}$\hat{f}_{std}$ (smoothed) & \cellcolor{smoothbg}88.5\% & \cellcolor{smoothbg}\textbf{58.5}\% \\
& & $f_{cbm}$ & \textbf{100}\% & 41\% \\
& & \cellcolor{smoothbg}$\hat{f}_{cbm}$ (smoothed) & \cellcolor{smoothbg}82.5\% & \cellcolor{smoothbg}\textbf{55.5}\% \\
\midrule

\multirow{8}{*}{0.01}
& \multirow{4}{*}{latent $z$}
& $f_{std}$& 100\% & 29.5\% \\
& & \cellcolor{smoothbg}$\hat{f}_{std}$ (smoothed) & \cellcolor{smoothbg}100\% & \cellcolor{smoothbg}\textbf{33.5}\% \\
& & $f_{cbm}$ & 100\% & 37.5\% \\
& & \cellcolor{smoothbg}$\hat{f}_{cbm}$ (smoothed) & \cellcolor{smoothbg}100\% & \cellcolor{smoothbg}\textbf{47}\% \\
& \multirow{4}{*}{concept $c$}
& $f_{std}$& 100\% & 72\% \\
& & \cellcolor{smoothbg}$\hat{f}_{std}$ (smoothed) & \cellcolor{smoothbg}88.5\% & \cellcolor{smoothbg}\textbf{80.5}\% \\
& & $f_{cbm}$ & \textbf{100}\% & 60.5\% \\
& & \cellcolor{smoothbg}$\hat{f}_{cbm}$ (smoothed) & \cellcolor{smoothbg}82.5\% & \cellcolor{smoothbg}\textbf{71}\% \\
\bottomrule
\end{tabular}
\end{table*}
\section{Cross-Family Evaluation Beyond CBMs}
\label{app:cross_family}

We extend the generator-based evaluation to PixPNet, a prototype-based
interpretable classifier, alongside ResNet and PCBM. The generator provides a
common source of perturbations, while each classifier processes the generated
images through its own architecture. This allows us to evaluate prediction
stability and randomized-smoothing certificates across model families without
matching generator concepts to PixPNet prototypes.

\subsection{Models and Evaluation Protocol}
\label{app:cross_family_protocol}

We consider a standard ResNet classifier, a PCBM with CAV-style concept
representations, and PixPNet with localized part prototypes. For a downstream
classifier $f_m$ and the fixed generator $g$, we evaluate the composite
mapping $F_m(z)=f_m(g(z))$. Gaussian noise is applied to the generator latent
code, whereas Bernoulli interventions flip binary concepts in the generator's
concept representation before decoding. In both cases, the perturbations are
defined independently of the downstream classifier architecture.

For the ResNet--PixPNet comparison, we retain 400 generated inputs on which
both models agree and use their shared clean prediction as the reference
label. This follows the agreement-set principle used in the main ResNet--PCBM
experiments, but produces a separate evaluation set. The cross-family
Bernoulli experiment uses $K=10$ binary generator concepts and reports pooled
statistics over these 400 inputs. By contrast, the main ResNet--PCBM
experiment uses $K=15$ concepts and reports statistics over five subsets of
200 images. The two Bernoulli evaluations therefore differ in both concept
dimensionality and sample aggregation, and their absolute SmAcc and certified
radius values should not be compared directly.

We measure prediction flip rate (PFR) for empirical stability and use
perturbation-specific randomized smoothing for certification. Smoothed
accuracy (SmAcc) is the fraction of non-abstaining predictions that match the
reference label, with abstentions counted as incorrect in the denominator.
Certified radii are averaged over samples that are correct relative to the
reference label and do not abstain.

\subsection{Empirical Prediction Stability}
\label{app:cross_family_empirical}

Table~\ref{tab:cross_family_pfr} reports PFR for ResNet and PixPNet under
Gaussian latent perturbations and Bernoulli generator-concept flips. PFR
increases with perturbation strength for both models. PixPNet exhibits a
higher flip rate than ResNet in all four reported settings, demonstrating that
prototype-based predictions can be analyzed under the same externally defined
perturbations.

\begin{table}[t]
\centering
\caption{Prediction flip rate (PFR) of ResNet and PixPNet under
 generator-defined perturbations on CUB-200-2011. Lower is better.}
\label{tab:cross_family_pfr}
\begin{tabular}{llrr}
\hline
Perturbation & Strength & ResNet PFR (\%) & PixPNet PFR (\%) \\
\hline
Gaussian latent & $\sigma=0.05$ & 23.99 & 28.71 \\
Gaussian latent & $\sigma=0.10$ & 39.21 & 44.03 \\
Bernoulli semantic & $\rho=0.10$ & 33.97 & 36.58 \\
Bernoulli semantic & $\rho=0.20$ & 54.31 & 57.40 \\
\hline
\end{tabular}
\end{table}

\subsection{Gaussian Randomized Smoothing}
\label{app:cross_family_gaussian}

We apply Gaussian smoothing to $F_m$ by sampling latent perturbations
$\delta\sim\mathcal{N}(0,\sigma^2 I)$ and evaluating $F_m(z+\delta)$.
The resulting certificate is an $\ell_2$ radius in generator latent space.
Table~\ref{tab:cross_family_gaussian} reports SmAcc and average certified
radius (ACR) for ResNet, PCBM, and PixPNet.

All three model families retain nonzero certified radii across the tested
noise scales. Increasing $\sigma$ generally reduces SmAcc, while the reported
average radius increases. The relative ordering changes across noise levels
and metrics: for example, PixPNet has higher SmAcc than ResNet at
$\sigma=0.05$ and $0.10$, but a smaller ACR at both noise levels. These
results illustrate why predictive accuracy and certified radius should be
reported separately.

\begin{table}[t]
\centering
\caption{Gaussian latent-space randomized smoothing on CUB-200-2011.
 SmAcc is reported in percent; ACR is the average certified latent
 $\ell_2$ radius among correct, non-abstaining predictions.}
\label{tab:cross_family_gaussian}

\begin{tabular}{crrrrrr}
\hline
& \multicolumn{2}{c}{ResNet} & \multicolumn{2}{c}{PCBM} & \multicolumn{2}{c}{PixPNet} \\
$\sigma$ & SmAcc & ACR & SmAcc & ACR & SmAcc & ACR \\
\hline
0.01 & 96.20 & 0.0228 & 97.40 & 0.0215 & 95.00 & 0.0199 \\
0.05 & 88.40 & 0.0747 & 84.90 & 0.0672 & 88.90 & 0.0553 \\
0.10 & 81.70 & 0.1034 & 77.20 & 0.0918 & 82.90 & 0.0682 \\
0.30 & 62.40 & 0.1263 & 57.10 & 0.1183 & 56.20 & 0.0839 \\
\hline
\end{tabular}%

\end{table}

\subsection{Bernoulli Generator-Concept Smoothing}
\label{app:cross_family_bernoulli}

We next apply independent Bernoulli flips to the $K=10$ binary generator
concepts, with flip probability $\rho$ per concept. The expected number of
flipped concepts is $10\rho$. Each intervened concept representation is
decoded into an image and classified by ResNet or PixPNet. Table~
\ref{tab:cross_family_bernoulli_smacc} reports smoothed accuracy on the
400-image ResNet--PixPNet agreement set.

SmAcc decreases as the intervention probability increases. At $\rho=0.05$,
the smoothed predictions agree with the reference labels for nearly every
image; at $\rho=0.30$, the corresponding fractions drop to 12.50\% for ResNet
and 9.00\% for PixPNet. These results describe prediction consistency under
increasing semantic randomization in the generator concept space.

\begin{table}[t]
\centering
\caption{Bernoulli generator-concept smoothing on the ResNet–PixPNet agreement set}
\label{tab:cross_family_bernoulli_smacc}
\begin{tabular}{rrrr}
\hline
$\rho$ & Expected flips & ResNet SmAcc (\%) & PixPNet SmAcc (\%) \\
\hline
0.05 & 0.5 & 100.00 & 99.75 \\
0.10 & 1.0 & 75.25 & 71.25 \\
0.20 & 2.0 & 28.75 & 23.75 \\
0.30 & 3.0 & 12.50 & 9.00 \\
\hline
\end{tabular}
\end{table}

\subsection{Certified Robustness to Concept Flips}
\label{app:cross_family_hamming}

For Bernoulli smoothing, we use the Hamming-space certification procedure
from Appendix~B.4 to certify a number of additional adversarial generator-
concept flips. Table~\ref{tab:cross_family_hamming} reports the average
certified Hamming radius (ACR-H) and certified accuracy at radius one
(CA@1). ACR-H is averaged over correct, non-abstaining inputs. CA@1 is the
percentage of all 400 inputs whose smoothed prediction matches the reference
label, does not abstain, and has certified radius at least one.

Certification is limited for both classifiers: CA@1 remains between 1.00\%
and 1.75\% in the reported settings. The larger ACR-H values at higher
$\rho$ do not indicate stronger overall accuracy, because ACR-H conditions on
the shrinking set of correct, non-abstaining predictions. Reporting CA@1
alongside ACR-H makes this distinction explicit.

\begin{table}[t]
\centering
\caption{Bernoulli concept-space certification on the 400-image
 ResNet--PixPNet agreement set. ACR-H is the mean certified Hamming radius
 over correct, non-abstaining predictions; CA@1 is computed over all
 400 images.}
\label{tab:cross_family_hamming}
\begin{tabular}{rrrrr}
\hline
& \multicolumn{2}{c}{ACR-H} & \multicolumn{2}{c}{CA@1 (\%)} \\
$\rho$ & ResNet & PixPNet & ResNet & PixPNet \\
\hline
0.05 & 0.0125 & 0.0150 & 1.25 & 1.50 \\
0.10 & 0.0133 & 0.0211 & 1.00 & 1.50 \\
0.20 & 0.0609 & 0.0526 & 1.50 & 1.25 \\
0.30 & 0.2200 & 0.2222 & 1.75 & 1.50 \\
\hline
\end{tabular}
\end{table}

\subsection{Discussion}
\label{app:cross_family_discussion}

The PixPNet results extend the evaluation beyond concept bottlenecks.
Across standard, concept-based, and prototype-based models, a fixed generator
provides the perturbations and the downstream classifier supplies predictions.
Gaussian latent noise yields an $\ell_2$ certificate, while Bernoulli
concept flips yield a Hamming-space certificate. The findings establish the
applicability of this evaluation procedure to different classifier families;
they do not imply a uniform robustness advantage for interpretable models.
Our detailed analysis of concept sensitivity and vocabulary design remains
focused on CBMs, for which the intermediate concept representation is
explicitly available.


\end{document}